%% file: main.tex
\documentclass[nohyperref]{article}

\usepackage[svgnames,table,dvipsnames]{xcolor}
\usepackage{microtype}
\usepackage{graphicx}
\usepackage{subfigure}
\usepackage{booktabs} 
\usepackage{hyperref}
\usepackage{url}

\usepackage[accepted]{icml2022}
\usepackage{amsmath}
\usepackage{amssymb}
\usepackage{mathtools}
\usepackage{amsthm}

\usepackage[capitalize,noabbrev]{cleveref}

\input{preamble/preamble}
\input{preamble/preamble_acronyms}

\input{preamble/preamble_math}
\input{math_commands.tex}

\usepackage[textsize=tiny]{todonotes}

\icmltitlerunning{Set Norm and Equivariant Skip Connections: Putting the Deep in Deep Sets}

\begin{document}
\interfootnotelinepenalty=10000

\twocolumn[
\icmltitle{Set Norm and Equivariant Skip Connections: Putting the Deep in Deep Sets}

\icmlsetsymbol{equal}{*}

\begin{icmlauthorlist}
\icmlauthor{Lily H. Zhang}{equal,nyu}
\icmlauthor{Veronica Tozzo}{equal,mgh,harvard}
\icmlauthor{John M. Higgins}{mgh,harvard}
\icmlauthor{Rajesh Ranganath}{nyu,nyucs}
\end{icmlauthorlist}

\icmlaffiliation{nyu}{Center for Data Science, New York University, New York, NY}
\icmlaffiliation{mgh}{Massachusetts General Hospital, Harvard Medical School, Cambridge, MA}
\icmlaffiliation{harvard}{Department of Systems Biology, Harvard Medical School, Boston, MA}
\icmlaffiliation{nyucs}{Department of Computer Science, New York University, New York, NY}

\icmlcorrespondingauthor{Lily H. Zhang}{lily.h.zhang@nyu.edu}
\icmlcorrespondingauthor{Veronica Tozzo}{vtozzo@mgh.harvard.edu}

\icmlkeywords{Machine Learning, ICML, deep learning}

\vskip 0.3in
]

\printAffiliationsAndNotice{\icmlEqualContribution} 

\begin{abstract}
Permutation invariant neural networks are a promising tool for making predictions from sets.
However, we show that existing permutation invariant architectures, Deep Sets and Set Transformer, can suffer from vanishing or exploding gradients
when they are deep.  
Additionally, layer norm, the normalization of choice in Set Transformer, can hurt performance by removing information useful for prediction. To address these issues, we introduce the ``clean path principle'' for equivariant residual connections and 
develop \gls{ipn}, a normalization tailored for sets. With these, we build 
Deep Sets++ and Set Transformer++,  models that reach high depths with better or comparable performance
than their original counterparts on a diverse suite of tasks.
We additionally introduce \gls{rbcdata}, a new single-cell dataset and real-world application of permutation invariant prediction.
We open-source our data and code here: \href{https://github.com/rajesh-lab/deep_permutation_invariant}{https://github.com/rajesh-lab/deep\_permutation\_invariant}.
\end{abstract}

\section{Introduction}

\input{ssections/intro}

\section{Permutation invariance}
\input{ssections/background}

\section{Problems with Existing Architectures}
\label{sec:issues}
\input{ssections/issues}

\section{Deep Sets++ and Set Transformer++}
\label{sec:solution}
\input{ssections/deeparch}

\section{FlowRBC}
\label{sec:flowRBC}
\input{ssections/flowrbc}

\section{Experimental setup}
\label{sec:experiments}

\input{ssections/experiments}

\section{Results}
\label{sec:results}
\input{ssections/results}
\section{Related Work}
\input{ssections/related}

\section{Conclusion}
\input{ssections/conclusions}

\section*{Acknowledgements}
This work was supported by
NIH/NHLBI Award R01HL148248, NSF Award 1922658
NRT-HDR: FUTURE Foundations, Translation, and Responsibility
for Data Science, a DeepMind Fellowship, and NIH R01 DK123330.

\bibliography{bibliography}
\bibliographystyle{icml2022}
\clearpage
\appendix

\input{appendix/hematocrit}

\input{appendix/layer_norm}
\input{appendix/norm_proofs}

\input{appendix/experiments}
\input{appendix/additional_results}
\input{appendix/ARC}
\input{appendix/residuals_gradients}

\end{document}

%% file: preamble/preamble.tex
\usepackage[acronym,shortcuts,smallcaps,nowarn,nohypertypes={acronym,notation}]{glossaries}

\usepackage{microtype}

\usepackage{booktabs} 
\usepackage{amsfonts}
\usepackage{amssymb}
\usepackage{amsmath}
\usepackage{amsthm}
\usepackage{enumitem}
\usepackage{soul}
\usepackage{bbm}
\usepackage{hyperref}
\usepackage{cleveref}
\usepackage{multirow}
\usepackage{rotating}
\usepackage{arydshln}
\usepackage{wrapfig}

\usepackage{pifont}

\usepackage{amssymb}

\definecolor{forestgreen}{RGB}{100,255,255}
\usepackage{afterpage}

\usepackage{multirow}

\usepackage{caption}
\usepackage{graphicx}
\usepackage{float}

\makeatletter
\renewcommand{\paragraph}{%
  \@startsection{paragraph}{4}%
  {\z@}{0ex \@plus .5ex \@minus .2ex}{-0.2em}%
  {\normalfont\normalsize\bfseries}%
}
\makeatother
\usepackage{makecell, cellspace, caption}

%% file: preamble/preamble_acronyms.tex
\newacronym{bsn}{bsn}{batch set norm}
\newacronym{fn}{fn}{feature norm}
\newacronym{bn}{bn}{batch norm}
\newacronym{ln}{ln}{layer norm}
\newacronym{in}{in}{instance norm}
\newacronym{ipn}{sn}{set norm}
\newacronym{msrc}{msrc}{moments-stabilizing residual connections}
\newacronym{rsn}{ResSetNet}{Residual Set Network}
\newacronym{rbcdata}{}{Flow-RBC}
\newacronym{erc}{erc}{equivariant residual connections}
\newacronym{ds++}{DS++}{Deep Sets++}
\newacronym{st++}{ST++}{Set Transformer++}

%% file: preamble/preamble_math.tex
\newcommand{\mbx}{\textbf{x}}
\newcommand{\mby}{\textbf{y}}
\newcommand{\mbz}{\textbf{z}}
\newcommand{\mba}{\textbf{a}}
\newcommand{\mbh}{\textbf{h}}
\newcommand{\mbp}{\textbf{p}}

\newtheorem{proposition}{Proposition}

%% file: math_commands.tex
\usepackage{amsmath,amsfonts,bm}

\def\eqref#1{equation~\ref{#1}}

\def\1{\bm{1}}

\DeclareMathAlphabet{\mathsfit}{\encodingdefault}{\sfdefault}{m}{sl}
\SetMathAlphabet{\mathsfit}{bold}{\encodingdefault}{\sfdefault}{bx}{n}



%% file: ssections/intro.tex
Many real-world tasks involve predictions on sets as inputs, from point cloud classification
\citep{Guo2020DeepLF, Wu20153DSA, Qi2017PointNetDL} to the prediction of health outcomes from single-cell data
\citep{regev2017science, Lhnemann2020ElevenGC, Liu2021MachineII, Yuan2017ChallengesAE}.

Models applied to input sets should satisfy
\emph{permutation invariance}: 
for any permutation of the elements in the input set, the model prediction stays the same. 
Deep Sets \citep{Zaheer2017DeepS} and Set Transformer \citep{Lee2019SetTA} are two general-purpose permutation-invariant neural networks that have been proven to be universal approximators of permutation-invariant functions under the right conditions \citep{Zaheer2017DeepS, Lee2019SetTA, Wagstaff2019OnTL}. 
In practice, however, 
these architectures are often tailored to specific tasks to achieve good performance 
\citep{Zaheer2017DeepS, Lee2019SetTA}.

In this work, we pursue a general approach to achieve improved performance: making permutation-invariant networks deeper. Whether deeper models benefit performance is often
task-dependent, but
the strategy of building deeper networks has yielded benefit for a variety of architectures and tasks \citep{he2016resenet, wang2019deeptransformers, Guohao2019deepGCN}. Motivated by these previous results, we investigate whether similar gains can be made of permutation-invariant architectures and prediction tasks on sets.

\begin{figure}
    \centering
    \subfigure[Deep Sets, Train Loss]{\includegraphics[width=40mm]{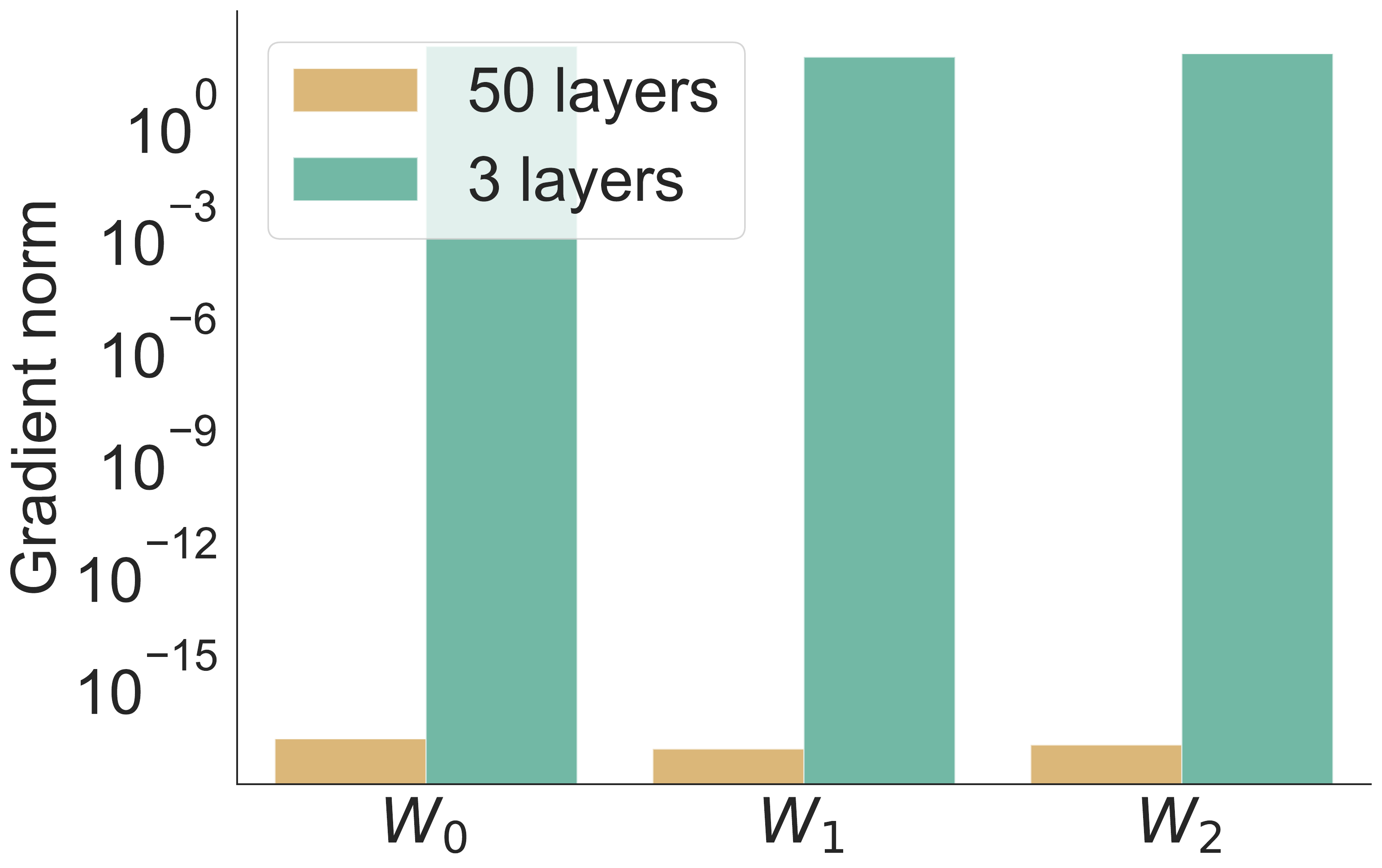}}
    \subfigure[Gradient Norms]{\includegraphics[width=40mm]{images2/ds_grads.pdf}}
    \subfigure[Set Transformer, Train Loss]{\includegraphics[width=40mm]{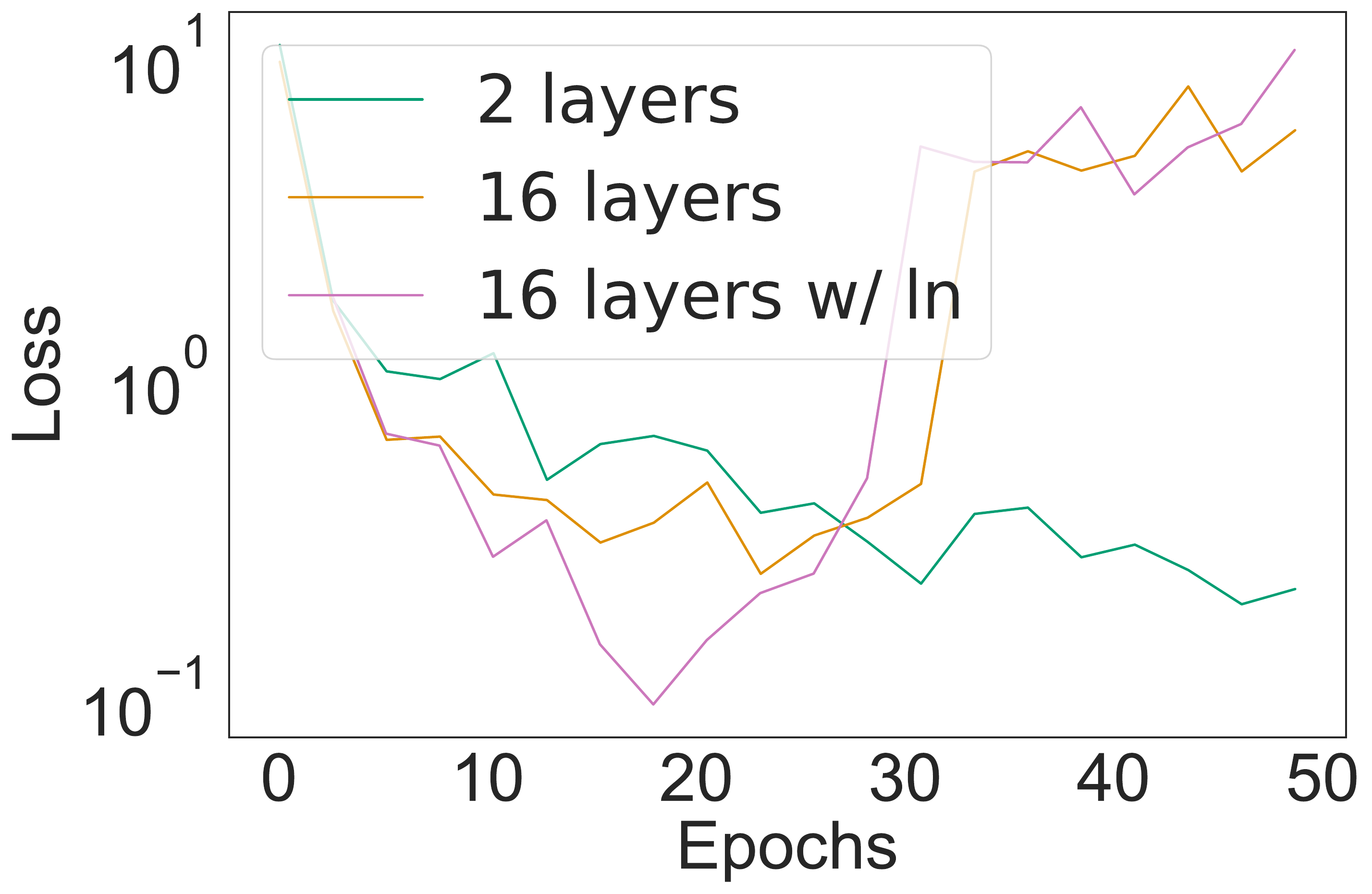}}
    \subfigure[Gradient Norms]{\includegraphics[width=40mm]{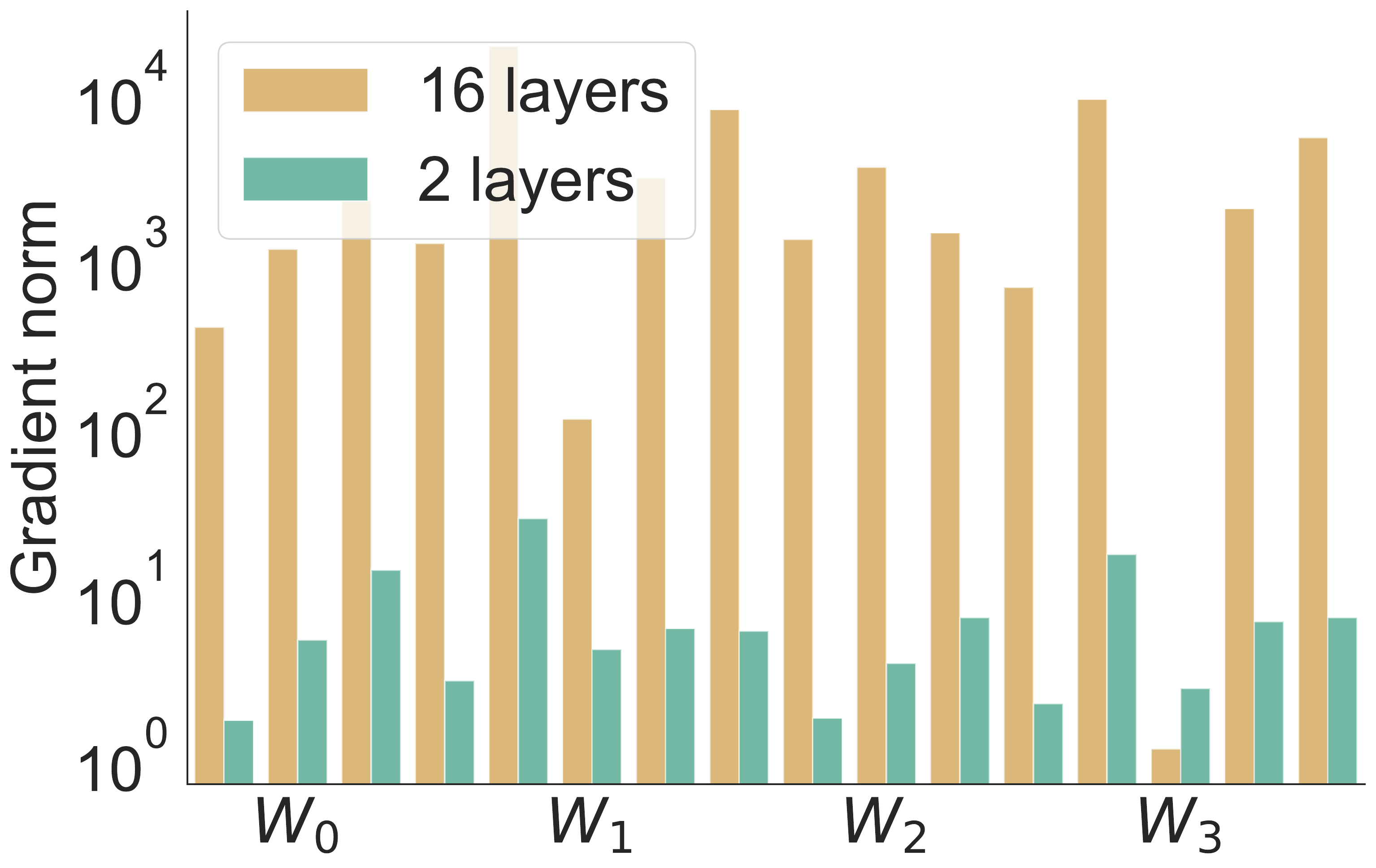}}
    \caption{At high depths, Deep Sets can suffer from vanishing gradients (top), while Set Transformer can suffer from exploding gradients (bottom). Experiment is MNIST digit variance prediction (see \Cref{sec:experiments} for details).}
    \label{fig:issues}
\end{figure}

However, naively increasing layers in Deep Sets and Set Transformer can hurt performance (see \Cref{fig:issues}). We show empirical evidence, supported by a gradient analysis, that both models can suffer from vanishing 
or exploding gradients (\Cref{sec:deepsets_vanishing}, \Cref{sec:settransformer_exploding}). 
Moreover, we observe that layer norm, the normalization layer discussed in Set Transformer, can actually hurt performance on tasks with real-valued sets, as its standardization forces potentially unwanted invariance to scalar transformations in set elements (\Cref{sec:layernorm_issues}).

To address these failures, we introduce Deep Sets++ and Set Transformer++, new versions of Deep Sets and Set Transformer with
carefully designed residual connections and normalization layers (\Cref{sec:solution}). 
First, we propose skip connections that adhere to what we call the ``clean path'' principle to address potential gradient issues.
Next, we propose 
set norm (\gls{ipn}), an easy-to-implement normalization layer for sets which standardizes each set over the minimal number of dimensions.
We consider both residual connections and normalization layers 
since
either alone can still suffer from gradient problems
\citep{zhang2018fixup, Yang2019AMF, De2020BatchNB}.

Deep Sets++ and Set Transformer++ 
are able to train at high depths 
without suffering from the issues seen in the original models (\Cref{sec:results}). Furthermore,
deep versions of these architectures improve upon their shallow counterparts on many tasks, avoiding issues such as exploding or vanishing gradients.
Among other results, these new architectures yield better accuracy on point cloud classification than the task-specific architectures proposed in the original Deep Sets and Set Transformer papers.

We also introduce a new dataset 
for permutation-invariant prediction called Flow-RBC (\Cref{sec:flowRBC}). The dataset consists of
red blood cell (RBC) measurements and hematocrit levels (i.e. the fraction of blood volume occupied by RBCs) for 100,000+ patients.  The size and presence of a prediction target (hematocrit) makes this dataset unique, even among single-cell datasets in established repositories like the Human Cell Atlas \citep{regev2017science}.  Given growing interest around single-cell data for biomedical science \citep{Lhnemann2020ElevenGC}, Flow-RBC provides machine learning researchers with the opportunity to benchmark their methods on an exciting new 
real-world application.

%% file: ssections/background.tex
Let $M$ be the number of elements in a set, and let $\mbx$ denote a single set with samples $\mbx_1, ..., \mbx_M, \mbx_i \in \mathcal{X}$. A function $f: \mathcal{X}^M \rightarrow \mathcal{Y}$ is \emph{permutation invariant} if any permutation $\pi$ of the input set results in the same output:
$f(\pi \mbx) = f(\mbx).$
A function $\mathbf{\sigma}: \mathcal{X}^M \rightarrow \mathcal{Y}^M$ is \emph{permutation equivariant} if, for any permutation $\pi$, the outputs are permuted accordingly:
$\sigma(\pi \mbx) = \pi \sigma(\mbx).$
A function is permutation-invariant if and only if it is sum-decomposable with sufficient conditions on the latent space dimension
\citep{Zaheer2017DeepS, Wagstaff2019OnTL}. 
A sum-decomposable function $f: \mathcal{X}^M \rightarrow \mathcal{Y}$ is one which can be expressed using a function $\phi: \mathcal{X} \rightarrow \mathcal{Z}$ mapping each input element to a latent vector,
a sum aggregation over the elements of the resulting output,
and an unconstrained decoder $\rho: \mathcal{Z} \rightarrow \mathcal{Y}$: 
\begin{equation}\label{eq:sum-decomposable}
     f(\mathbf{x}) = \rho\left(\sum_{i=1}^M\phi(\mbx_i)\right).
\end{equation}

Existing permutation-invariant architectures utilize the above fact to motivate their architectures, which consist of an equivariant encoder, permutation-invariant aggregation, and unrestricted decoder. Equivariant encoders can express $\phi(\mbx_i)$ for each element $\mbx_i$ if interactions between elements are zeroed out. For the remainder of the paper, we 
consider the depth of a permutation-invariant network to be the number of layers in the equivariant encoder. We do not consider decoder changes as the decoder is any unconstrained network, so we expect existing work on increasing depth to directly transfer.

%% file: ssections/issues.tex
Both Deep Sets and Set Transformer 
are permutation invariant \citep{Zaheer2017DeepS, Lee2019SetTA}.
However, a gradient analysis of each shows that both architectures can exhibit vanishing or exploding gradients. We present experimental evidence of vanishing and exploding gradients in Deep Sets and Set Transformer respectively (\Cref{fig:issues}).

\subsection{Deep Sets gradient analysis}
\label{sec:deepsets_vanishing}
Deep Sets consists of an encoder of equivariant feedforward layers  (where each layer is applied independently to each element in the set),
 a sum or max aggregation, and a decoder also made up of feedforward layers \citep{Zaheer2017DeepS}. Each feedforward layer is an affine transform with a ReLU non-linearity: for layer $\ell$ and element $i$, we have
$\mbz_{\ell, i} = \text{relu}(\mbz_{\ell - 1, i}W_\ell + b_\ell)$. 
We denote the output after an $L$-layer encoder and permutation-invariant sum aggregation as $\mby = \sum_i \mbz_{L, i}$.
Then, the gradient of weight matrix $W_1$ of the first layer is as follows:
\begin{align}
    \frac{\partial{\mathcal{L}}}{\partial{W_1}} &= \frac{\partial{\mathcal{L}}}{\partial{\mby}} \sum_i \frac{\partial{\mby}}{\partial{\mbz_{L, i}}}
     \frac{\partial{\mbz_{L, i}}}{\partial{W_1}}.
\end{align}

The rightmost term above is a product of terms which can become vanishingly small when the number of layers $L$ is large:
\begin{align}
\frac{\partial{\mbz_{L, i}}}{\partial{W_1}} &= \frac{\partial{\mbz_{L, i}}}{\partial{\mbz_{1, i}}}\frac{\partial{\mbz_{1, i}}}{\partial{W_1}}\\
    &=
    \prod_{\ell = 2}^L \frac{\partial{\mbz_{\ell, i}}}{\partial{\mbz_{\ell - 1, i}}}\frac{\partial{\mbz_{1, i}}}{\partial{W_1}}\\
    &= \frac{\partial{\mbz_{1, i}}}{\partial{W_1}}\prod_{\ell = 2}^L \frac{\partial{\text{relu}(\mbz_{\ell,i })}}{\partial{\mbz_{\ell, i}}}W_\ell.
\end{align}

This gradient calculation mirrors that of a vanilla feedforward network, except for the additional summation over each of the elements (or the corresponding operation for max aggregation). Despite the presence of the sum, the effect of a product over many layers of weights still dominates the overall effect on the gradient of earlier weights. 
We provide experimental evidence in  \Cref{fig:issues}.

\subsection{Set Transformer gradient analysis}
\label{sec:settransformer_exploding}
Set Transformer consists of an encoder, aggregation, and decoder built upon a multihead attention block (MAB) \citep{Lee2019SetTA}.\footnote{The MAB module in \citet{Lee2019SetTA} should not be confused with multihead attention \citep{Vaswani2017AttentionIA}, which is a component of the module.} The MAB differs from a transformer block in that its skip connection starts at the linearly transformed input $\mbx W_Q$ rather than $\mbx$ (see \Cref{eq:transformer}).\footnote{The implementation of the MAB module in code differs from the definition in the paper. We follow the former.} Let $\text{Attn}_K$ be multihead attention with $K$ heads and a scaled softmax, \textit{i.e.} $\text{softmax}(\cdot / \sqrt{D})$ where $D$ is the number of features. Then, MAB can be written as:
\begin{align}
    \text{MAB}_K(\mbx, \mby) &= f(\mbx, \mby) + \text{relu}(f(\mbx, \mby) W + \mathbf{b}), \label{eq:ff}\\
    f(\mbx, \mby) &= \mbx W_Q + \text{Attn}_K(\mbx, \mby, \mby). \label{eq:transformer}
\end{align}
The Set Transformer encoder block is a sequence of two MAB blocks, the first between learned inducing points and the input $\mbx$, and the second between the input $\mbx$ and the output of the first block. Given $D$ hidden units and $M$ learned inducing points $\mbp$,\footnote{Typically, $M << S$ for computational efficiency.} the inducing point set attention block (ISAB) can be written as such:
\begin{align}
    \text{ISAB}_M(\mbx) &= \text{MAB}_K(\mbx, \mbh) \in \mathbb{R}^{S\times D} \\
    \text{where } \mbh &= \text{MAB}_K(\mbp, \mbx) \in \mathbb{R}^{M \times D}. \label{eq:first_mab}
\end{align}
The aggregation block is an MAB module between a single inducing point and the output of the previous block ($M=1$ in \Cref{eq:first_mab}), and the decoder blocks are self-attention modules between the previous output and itself.
In \citet{Lee2019SetTA}, layer norm is applied to the outputs of \Cref{eq:ff} and \Cref{eq:transformer} in the MAB module definition but is turned off in the experiments.

Consider a single ISAB module. We let $\mbz_1$ denote the output of the previous block, $\mbz_2$ denote the output after the first MAB module (i.e. $\mbh$ in \Cref{eq:first_mab}), and $\mbz_3$ denote the output of the second MAB module, or the overall output of the ISAB module. Then,
\begin{align}
    f_1 &= f(\mathbf{p}, \mbz_1) = I W_{1Q} + \text{Attn}_K(\mathbf{p}, \mbz_1, \mbz_1) \\
    \mbz_2 &= f_1 + \text{relu}(f_1 W_1 + \mathbf{b}_1) \\
    f_2 &= f(\mbz_1, \mbz_2) = \mbz_1 W_{2Q} + \text{Attn}_K(\mbz_1, \mbz_2, \mbz_2) \\
    \mbz_3 &= f_2 + \text{relu}(f_2 W_2 + \mathbf{b}_2).
\end{align}
Let $I$ denote the identity matrix
Then, the gradient of a single ISAB block output $\mbz_3$ with respect to its input $\mbz_1$ can be represented as $\frac{\partial{\mbz_3}}{\partial{\mbz_1}} = \frac{\partial{\mbz_3}}{\partial{f_2}}\frac{\partial{f_2}}{\partial{\mbz_1}}$, or
\begin{align*}
    \Big(I + \frac{\partial{\text{relu}(f_2 W_2 + \mathbf{b}_2)}}{\partial{(f_2 W_2 + \mathbf{b}_2)}}W_2\Big)
    \Big(
    W_{2Q}+ \frac{\partial{\text{Attn}_K(\mbz_1, \mbz_2, \mbz_2)}}{\partial{\mbz_1}}\Big).
\end{align*}

In particular, we notice that even if the elements in 
$\frac{\partial{\text{relu}(f_2 W_2 + \mathbf{b}_2)}}{\partial{(f_2 W_2 + \mathbf{b}_2)}}W_2$ 
and $\frac{\partial{\text{Attn}_K(\mbz_1, \mbz_2, \mbz_2)}}{\partial{\mbz_1}}$ are close to zero, 
the weights $W_{2Q}$ will affect the partial derivatives of each ISAB output with respect to its input. 
The gradient of earlier weights will be the product of many terms of the above form, and
this product can explode when the magnitude of the weights grows, causing exploding gradients and unstable training (see \Cref{fig:issues}(c) for an example). We find experimentally that even with the addition of layer norm, the problem persists. See \Cref{sec:st_layernorm_ga} for an analogous gradient analysis with the inclusion of layer norm.

Based on the gradient analysis provided for both Deep Sets and Set Transformer, both vanishing and exploding gradients are possible for both models. In our experiments, we primarily see evidence of vanishing gradients for Deep Sets and exploding gradients for Set Transformer.

\subsection{Layer norm can hurt performance}
\label{sec:layernorm_issues}
\begin{table}[]
    \centering
    \begin{tabular}{lrr}
    \toprule
       & No norm & Layer norm \\
     \midrule
       Hematocrit & 18.7436 $\pm$ 0.0148 & \underline{19.0904 $\pm$ 0.1003}\\
       Point Cloud & 0.9217 $\pm$ 0.0119 & 0.9219 $\pm$ 0.0052\\
    Normal Var & 0.0023 $\pm$ 0.0006 & \underline{0.0801 $\pm$ 0.0076}\\
        \bottomrule
    \end{tabular}
    \caption{
   Set Transformer can perform worse (underlined) with layer norm than with no normalization, particularly when inputs are real-valued. 
    Results are test loss over three seeds (CE for Point Cloud, MSE for rest). Lower is better.}
    \label{tab:layer_norm}
\end{table}
Layer norm \citep{Ba2016LayerN} was introduced for permutation-invariant prediction tasks in Set Transformer \citep{Lee2019SetTA}, mirroring transformer architectures for other tasks. However, while layer norm has been shown to benefit performance in other settings \citep{Ba2016LayerN,Chen2018TheBO}, we find that layer norm can in fact hurt performance on certain tasks involving sets (see \Cref{tab:layer_norm}).

Let $\overrightarrow{\mu}_\mathbf{z}, \overrightarrow{\sigma}_\mathbf{z} \in \mathbb{R}^{D}$ be the statistics used for standardization of a vector $\mathbf{z} \in \mathbb{R}^D$ and $\overrightarrow{\gamma}, \overrightarrow{\beta} \in \mathbb{R}^D$ be transformation parameters acting on each feature independently. Then, given a set with elements $\{\mathbf{x}_i\}_{i=1}^M \in \mathbb{R}^D$, layer norm first standardizes each element independently 
$\bar{\mathbf{x}}_i = \frac{\mathbf{x}_i - \mu_{\mathbf{x}_i}}{\sigma_{\mathbf{x}_i}}, $
and then transforms
$ \hat{\mathbf{x}}_i = \mathbf{x}_i \odot \overrightarrow{\gamma} + \overrightarrow{\beta}.$

Element-wise standardization forces an invariance where 
two elements whose activations differ in only a scale yield the same output when processed through layer norm following a linear projection.
If we consider layer norm in is typical placement, after a linear projection and before the non-linear activation $f(\mbx_i) = \text{relu}(\text{LN}(\mbx_i W)) $ \citep{Ba2016LayerN, Ioffe2015BatchNA, Ulyanov2016InstanceNT, Cai2021GraphNormAP}, 
we have that for $\mbx_i$ and $\mbx_{i'} = \alpha \mbx_i$, $\alpha \in \mathbb{R}$,

\begin{align}
    \text{LN}(\mbx_{i'} W)
    &= \frac{(\alpha \mbx_i ) W - \overrightarrow{\mu}_{\mbx_{i'} W}}{\overrightarrow{\sigma}_{\mbx_{i'} W}} * \overrightarrow{\gamma}+ \overrightarrow{\beta} \\
    &= \frac{\alpha\mbx_i W - \alpha \overrightarrow{\mu}_{\mbx_i W} }{\alpha \overrightarrow{\sigma}_{\mbx_i W} W } * \overrightarrow{\gamma}+ \overrightarrow{\beta} \\
    &= \frac{\mbx_i W - \overrightarrow{\mu}_{\mbx_i W}}{\overrightarrow{\sigma}_{\mbx_i W}} * \overrightarrow{\gamma}+ \overrightarrow{\beta} = \text{LN}(\mbx_i W).
\end{align}

Since $\text{LN}(\mbx_{i'} W) = \text{LN}(\mbx_{i} W)$, $f(\mbx_{i'}) = f(\mbx_i)$, meaning the two elements are indistinguishable at this point in the network. This invariance reduces representation power (two such samples cannot be treated differently in the learned function) and removes information which may potentially be useful for prediction (i.e. per-element mean and standard deviation).

\paragraph{An Example in 2D.} 
Consider
sets of two-dimensional real-valued elements and a model with 2D activations.
Layer norm's standardization will map all elements 
to either (-1, 1), (0, 0), or (1, -1),
corresponding to whether 
the first coordinate of each element is less than, greater than, or equal to the second coordinate. 
If the task is classifying 2D point clouds,
any two shapes which share the same division of points on either side of the $y=x$ line will be indistinguishable (see \Cref{sec:2d_ln} for a visualization). 
Generalizing this phenomenon to higher dimensions, layer norm's standardization decreases the degrees of freedom in elements' outputs relative to their inputs, an effect that can be particularly harmful for sets of low-dimensional, real-valued elements.
In contrast, layer norm is commonly used in NLP, where
one-hot encoded categorical tokens 
will not be immediately mapped to the same outputs. 
Differences such as these ones highlight the need to consider normalization layers tailored to the task and data type at hand. 

Our analysis on gradients and layer norm 
does not suggest that these issues 
will always be present.
However, the possibility of these issues, as well as experimental evidence thereof, raises the need 
for alternatives which do not exhibit the same problems.

%% file: ssections/deeparch.tex
We propose \glsfirst{ds++} and \glsfirst{st++},
new architectures  that differ from the originals only in their encoders, as 
we fix the decoder and aggregation
to their original versions. 
For simplicity, we let the hidden dimension remain constant throughout the encoder. 
Based on the analysis of \Cref{sec:issues}, 
we explore alternative residual connections scheme to fix the vanishing and exploding gradients. 
Moreover, given the potential issues with layer norm for real-valued set inputs, we consider an alternative normalization.
Concretely, we propose 
the clean-path equivariant residual connections 
and set norm.

\subsection{Clean-path equivariant residual connections}

 \begin{figure*}[t]
\centering
\subfigure[Non-clean path for Set Transformer\label{fig:greypathpostnorm} ]{\includegraphics[height=50mm]{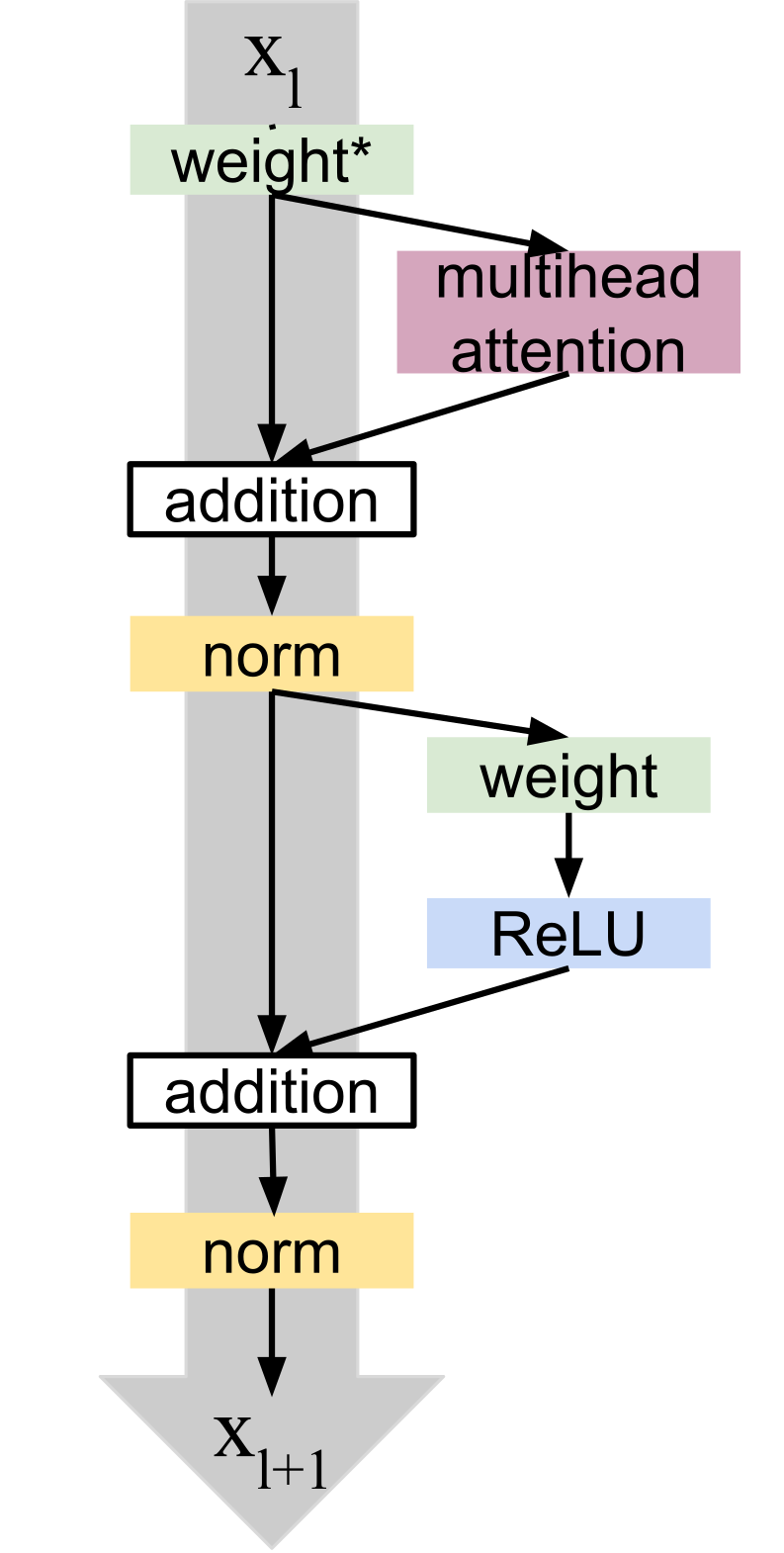}}
      \hspace{1cm}
      \subfigure[Clean path for Set Transformer\label{fig:greypathprenorm} ]{\includegraphics[height=50mm]{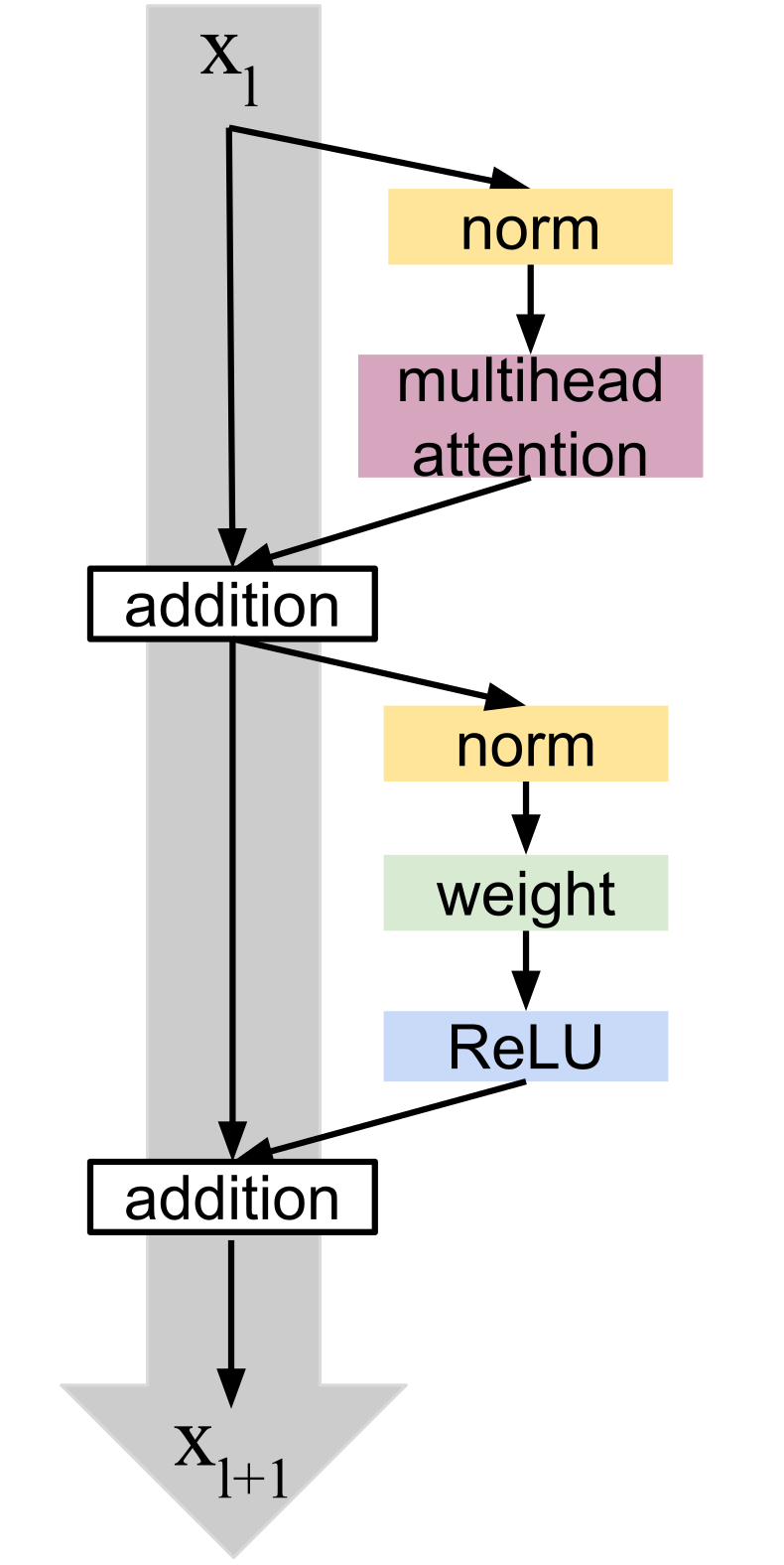}}
           \hspace{1cm}
 \subfigure[Non-clean path for Deep Sets\label{fig:greypathresnet} ]{\includegraphics[height=50mm]{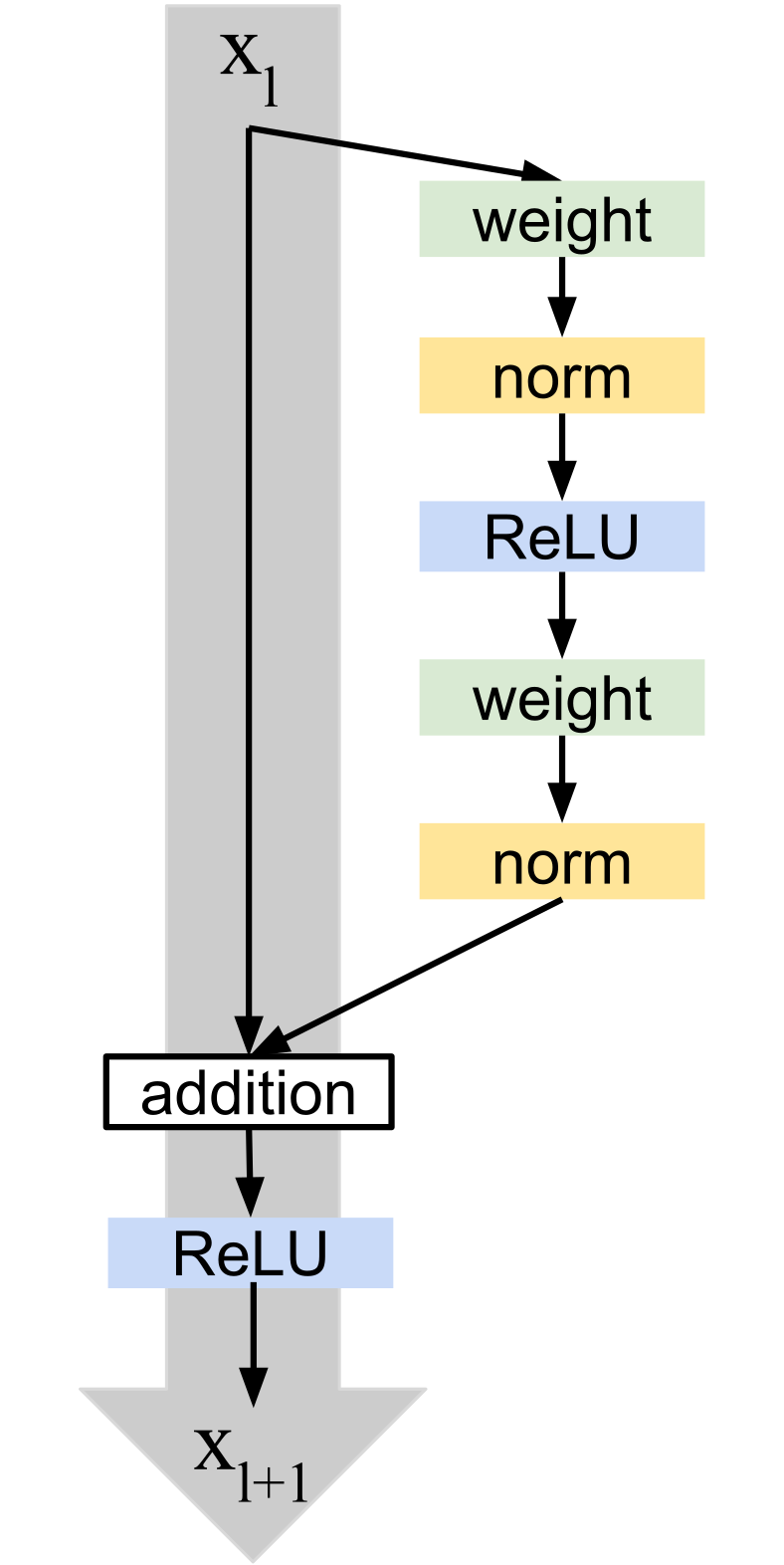}}
 \hspace{1cm}
    \subfigure[Clean path for Deep Sets\label{fig:greypathhe} ]{\includegraphics[height=50mm]{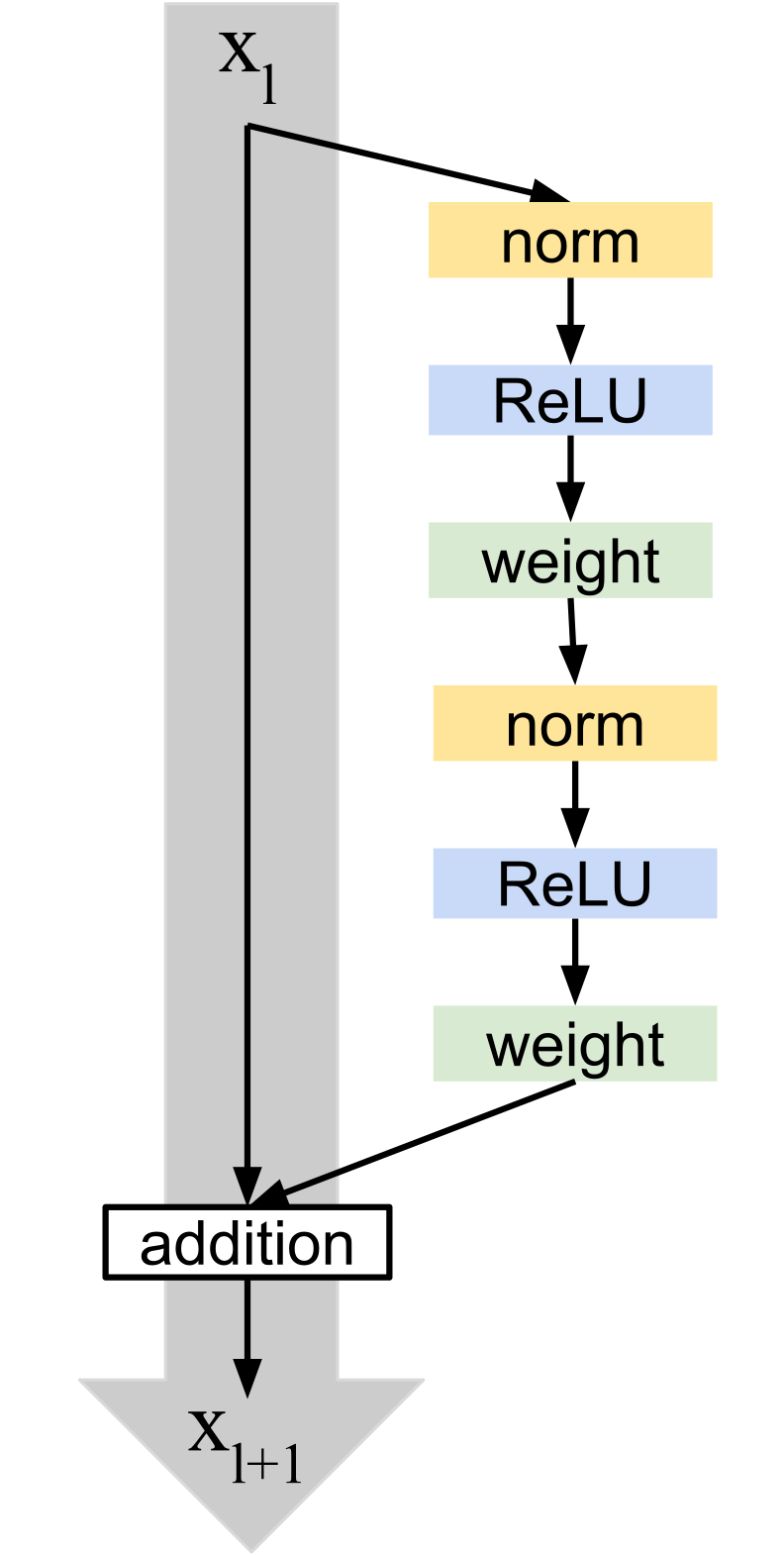}}
    \caption{Clean path variants have no additional operations on the residual path (denoted by a grey arrow), whereas non-clean path variants do.
    In (c), weight* is also part of the attention computation.}
    \label{fig:graypath}
\end{figure*}
Let $f$ be an equivariant function where $\mathcal{X} = \mathcal{Y} = \mathbb{R}^D$, i.e. $f: \mathbb{R}^{M \times D} \rightarrow \mathbb{R}^{M \times D}$. A function $g$ which adds 
each input to its output after applying any equivariant function $f$ is also equivariant:

\begin{equation*}
g(\pi \mbx) = f(\pi \mbx) + \pi \mbx 
= \pi f(\mbx) +  \pi \mbx 
= \pi g(\mbx).
\end{equation*}

While such residual connections exist in the literature \citep{weiler2019erc, wang2020erc}, here we refer to them as \emph{equivariant residual connections} (ERC) to highlight their equivariant property and differentiate them from other possible connections that skip over blocks (see \Cref{sec:results} for an example).
In sets, ERCs act on every element
and
eliminate the vanishing gradient problem (see \Cref{sec:residual_gradients} for a gradient analysis).

ERCs can be placed in different arrangements within an architecture
\citep{he2016resenet, He2016IdentityMI, Vaswani2017AttentionIA, Vaswani2018Tensor2TensorFN}.
We consider \emph{non-clean path} and \emph{clean path} arrangements. Let $l$ indicate the layer in the network.
Non-clean path blocks
include operations before or after the residual connections and must be expressed as either
\begin{equation}
    \mbx_{l+1} = g(\mbx_l) + f(\mbx_l) \quad \text{ or } \quad \mbx_{l+1} = g(\mbx_l + f(\mbx_l)),
\end{equation}
where $g, f$ cannot be the identity function. This arrangement was used in the MAB module of the Set Transformer architecture (see \Cref{fig:graypath} panel a). 
Previous literature on non permutation-invariant architectures shows that the presence of certain operations between skip connections could yield undesirable effects \citep{He2016IdentityMI, he2016resenet, Klein2017OpenNMTOT, Vaswani2018Tensor2TensorFN, Xiong2020OnLN}.

In contrast, \emph{clean path} arrangements add the unmodified input to a function applied on it,
\begin{equation}
    \mbx_{l+1} = \mbx_l + f(\mbx_l),
\end{equation}
resulting in a clean path from input to output (see gray arrows in \Cref{fig:graypath} b and d).
The clean path MAB block (\Cref{fig:graypath} panel b) mirrors the operation order of the Pre-LN Transformer \citep{Klein2017OpenNMTOT, Vaswani2018Tensor2TensorFN}, while the clean path version of Deep Sets mirrors that of the modified ResNet in \citet{He2016IdentityMI} (\Cref{fig:graypath} panel d).

\subsection{Set norm}
\label{sec:setnorm}
Designing normalization layers for permutation equivariant encoders 
requires careful consideration, as not all normalization layers are appropriate to use. To this aim, we analyze normalization layers 
as a composition of two operations: standardization and transformation. This setting captures most common normalizations \citep{Ioffe2015BatchNA, Ba2016LayerN, Ulyanov2016InstanceNT}.

Let $\mba \in \mathbb{R}^{N \times M \times D}$ be the activation before the normalization operation, where $N$ is the size of the batch,
$M$ is the number of elements in a set (sets are zero-padded to the largest set size),
and $D$ is the feature dimension.
First, the activations are standardized based on a setting $\mathcal{S}$ which defines which dimensions utilize separate statistics. For instance, $\mathcal{S} = \{N, M\}$ denotes that each set in a batch and each element in a set calculates its own mean and standard deviation for standardization, e.g. $\mu_{\mathcal{S}}(\mba)_{b,s} = \frac{1}{D}\sum_{d=1}^D \mba_{n, i, d}$. 
Results are repeated over the dimensions not in $\mathcal{S}$ so that $\mu_\mathcal{S}(\mba), \sigma_\mathcal{S}(\mba) \in \mathbb{R}^{N \times M \times D}$ match $\mba$ in dimensions for elementwise subtraction and division.
A standardization operation can be defined as:
\begin{equation} \label{eq:standardize}
    \bar\mba_\mathcal{S} = \frac{\mba - \mu_\mathcal{S}(\mba)}{\mathbf{\sigma}_\mathcal{S}(\mba)},
\end{equation}
where we assume that the division is well-defined (i.e. non-zero standard deviation). 

Next, the standardized activations are transformed through learned parameters which differ only over a setting of dimensions $\mathcal{T}$. For instance, $\mathcal{T} = \{D\}$ denotes that each feature is transformed by a different scale and bias, which are shared across the sets in the batch and elements in the sets.  Let $\overrightarrow{\gamma}_\mathcal{T}, \overrightarrow{\beta}_\mathcal{T} \in \mathbb{R}^{N \times M \times D}$ denote the learned parameters
and $\odot$ represent elementwise multiplication. Any transformation operation can be defined as:
\begin{equation} \label{eq:transform}
    \hat{\mba}_\mathcal{T} = \bar\mba \odot \overrightarrow{\gamma}_\mathcal{T} + \overrightarrow{\beta}_\mathcal{T}.
\end{equation}

\begin{proposition}{ 
Let $\mathcal{F}$ be the family of transformation functions which can be expressed via \Cref{eq:transform}.
Then, for $f \in \mathcal{F}$, $\mathcal{T} = \{D\}$ and $\mathcal{T}=\{\}$ are the only settings satisfying the following properties: 
\begin{enumerate}
    \item $f_\mathcal{T}(\pi_i \mba) = \pi_i f_\mathcal{T}(\mba)$ where $\pi_i$ is a permutation function that operates on elements in a set; 
    \item $f_\mathcal{T}(\pi_n \mba) = \pi_n f_\mathcal{T}(\mba)$ where $\pi_n$ is a permutation function that operates on sets.
\end{enumerate}} 
\end{proposition}
 See \Cref{sec:norm_proofs} for proof. 
In simpler terms, the settings  $\mathcal{T} = \{D\}$ and $\mathcal{T}=\{\}$ are the only ones that maintain permutation invariance and are agnostic to set position in the batch. The setting $\mathcal{T} = \{D\}$
contains $\mathcal{T}=\{\}$ and is more expressive, as $\mathcal{T} = \{\}$ is equivalent to $\mathcal{T} = \{D\}$ where learned parameters $\overrightarrow{\gamma}_\mathcal{T}, \overrightarrow{\beta}_\mathcal{T}$ each consist of a single unique value. Thus, we choose $\mathcal{T} = \{D\}$ as our choice of transformation.

Standardization will always remove information; certain mean and variance information become unrecoverable.
However, it is possible to control what information is lost based on the choice of dimensions over which standardization occurs.

With this in mind, we propose \emph{set norm} (\gls{ipn}), a new normalization layer designed to standardize over the fewest number of dimensions
of any standardization which acts on each set separately. 
Per-set standardizations are a more practical option for sets than standardizations which happen over a batch ($N \not\in \mathcal{S}$, batch norm is an example), as the latter introduce issues such as inducing dependence between inputs, requiring different procedures during train and test, and needing tricks such as running statistics to be stable. In addition, any standardization over a batch needs to take into account
how to weight differentially-sized sets in calculating the statistics as well as how to deal with small batch sizes caused by large inputs.

Set norm is a normalization defined by a per set standardization and per feature transformation ($\mathcal{L} = \{N\}, \mathcal{T} = \{D\}$): 
\begin{align*}
   \text{SN}(a_{nid}) &= \frac{\mba_{n} - \mu_n}{\sigma_n} \odot \gamma_{d} + \beta_d,\\  \mu_n &= \frac{1}{M}\frac{1}{D}\sum_{i=1}^M\sum_{d=1}^D a_{nid} , \\\sigma^2_n &= \frac{1}{M}\frac{1}{D}\sum_{i=1}^M\sum_{d=1}^D (a_{nid} - \mu_n)^2 .
\end{align*}
Set norm is permutation equivariant (see \Cref{sec:norm_proofs} for proof). It also standardizes over the fewest dimensions possible of any per-set standardization, resulting in the least amount of mean and variance information removed (e.g. only the global mean and variance of the set rather than the mean and variance of each sample in the set in the case of layer norm). 
Note that set norm assumes sets of size greater than one ($M>1$) or multi-sets in which at least two elements are different. 

\begin{table}[h]
    \centering  
        \centering
    \caption{Set norm can improve performance of 50-layer Deep Sets, while layer norm does not (Point Cloud, MNIST Var). In some cases, normalization alone is not enough to overcome vanishing gradients (Hematocrit, Normal Var). Table reports test loss (CE for Point Cloud, MSE otherwise). Lower is better. Uunderlined results are notable failures.}\label{tab:ds_norm}
         \centering
   \resizebox{0.5\textwidth}{!}{ \begin{tabular}{lrrr}
    \toprule
    & no norm & layer norm & set norm \\
    \midrule
    Hematocrit & 25.879 $\pm$ 0.001 & 25.875 $\pm$ 0.002 & 25.875 $\pm$ 0.002 \\
    Point Cloud & 3.609 $\pm$ 0.000 & \underline{3.619 $\pm$ 0.000}  & \textbf{1.542 $\pm$ 0.086} \\
    MNIST Var & 5.555 $\pm$ 0.001 & \underline{5.565 $\pm$ 0.001} & \textbf{0.259 $\pm$ 0.003} \\
     Normal Var & 8.4501 $\pm$ 0.0031 & 8.4498 $\pm$ 0.0054 & 8.4433 $\pm$ 0.0011 \\
        \bottomrule
    \end{tabular}}
 \end{table}

Next, we combine clean-path equivariant residual connections and set norm to build modified permutation-invariant architectures Deep Sets++ and Set Transformer++.

 \begin{table*}
        \centering
    \caption{Clean path residual connections outperform non-clean path residual connections both in Deep Sets and Set Transformer. Clean path residuals with set norm perform best overall. Results are test loss for deep architectures (50 layers Deep Set, 16 layers Set Transformer), lower is better.}\label{tab:clean_nonclean}
          \resizebox{\textwidth}{!}{ \begin{tabular}{lllcccc}
       \toprule
         Path & Residual type & Norm & Hematocrit (MSE) & Point Cloud (CE) & Mnist Var (MSE) & Normal Var (MSE) \\
         \hline
          Deep Sets & non-clean path &layer norm &19.6649 $\pm$ 0.0394 &\textbf{0.5974 $\pm$ 0.0022} & 0.3528 $\pm$ 0.0063 &
          1.4658 $\pm$ 0.7259 \\
         && feature norm & 19.9801 $\pm$ 0.0862 &0.6541 $\pm$ 0.0022 &\textbf{0.3371 $\pm$ 0.0059}
         & 0.8352 $\pm$ 0.3886 \\
         && set norm & 19.3146 $\pm$ 0.0409 &\textbf {0.6055 $\pm$ 0.0007} &\textbf{ 0.3421 $\pm$ 0.0022}& 
        0.2094 $\pm$ 0.1115 \\
           & clean path  & layer norm & 19.4192 $\pm$ 0.0173
 & 0.63682$\pm$ 0.0067 & 0.3997 $\pm$ 0.0302 & 
        0.0384 $\pm$ 0.0105 \\
         && feature norm &19.3917 $\pm$ 0.0685 &0.7148 $\pm$ 0.0164 &\textbf{0.3368 $\pm$0.0049} &
        0.1195 $\pm$ 0.0000\\
        \rowcolor{lightgray}
         && set norm &\textbf{19.2118 $\pm $ 0.0762} &0.7096 $\pm$ 0.0049 &\textbf{ 0.3441 $\pm $0.0036}
    & \textbf{0.0198 $\pm$ 0.0041} \\
\midrule
Set Transformer & non-clean path & layer norm & 19.1975 $\pm$ 0.1395 & 0.9219 $\pm$ 0.0052 & 2.0663 $\pm$ 1.0039 & 
0.0801 $\pm$ 0.0076\\
        && feature norm & 19.4968 $\pm$ 0.1442 & 0.8251 $\pm$0.0025 & \textbf{0.4043 $\pm$ 0.0078}
        & 0.0691 $\pm$ 0.0146 \\
        && set norm & 19.0521 $\pm $0.0288 & 1.9167 $\pm$ 0.4880 &\textbf{0.4064 $\pm$ 0.0147} &
        0.0249 $\pm$ 0.0112 \\
         &clean path 
       &  layer norm & \textbf{18.5747 $\pm$ 0.0263} & 0.6656 $\pm$ 0.0148 & 0.6383 $\pm$ 0.0020 &
    0.0104 $\pm$ 0.0000 \\
        && feature norm & 19.1967$\pm$ 0.0330 & \textbf{0.6188 $\pm$ 0.0141} & 0.7946 $\pm$0.0065 
        & 0.0074 $\pm$ 0.0010 \\
       \rowcolor{lightgray}
        && set norm & 18.7008 $\pm$ 0.0183  & \textbf{0.6280 $\pm $ 0.0098} & 0.8023 $\pm$ 0.0038 
        & \textbf{0.0030 $\pm$ 0.0000} \\
\bottomrule
    \end{tabular}}
\end{table*}

\subsection{\gls{ds++}}
\gls{ds++} adopts the building blocks mentioned above, 
resulting in a residual block of the form 
$$
\mbx_{l+1} = \mbx_{l} + \text{SetNorm}(W_{l1}(\text{relu}(\text{SetNorm}(W_{l2} \mbx_{l})))).
$$
The \gls{ds++} encoder starts with a first linear layer and no bias, as is customary before a normalization layer \citep{Ioffe2015BatchNA, Ba2016LayerN} and ends with a normalization-relu-weight operation after the final residual block in the encoder, following \citet{He2016IdentityMI}.

\subsection{\gls{st++}}
Similarly, \gls{st++} adds a set norm layer and adheres to the clean path principle (see \Cref{fig:graypath} (b)).
In practice, we define a variant of the ISAB model, which we call ISAB++, that changes the residual connections and adds normalization off the residual path, analogous to  the Pre-LN transformer \citep{Klein2017OpenNMTOT, Vaswani2018Tensor2TensorFN, Xiong2020OnLN}. 
We define two multi head attention blocks MAB$^1$ and MAB$^2$ with $K$ heads as 
\begin{align}
    &\text{MAB}^1_K(\mbx, \mby) = \mbh + \text{fcc}(\text{relu}(\text{SetNorm}(\mbh)))\\
    &\text{where } \mbh  = \mbx + \text{Attn}_K(\mbx, \text{SetNorm}(\mby), \mby).
\end{align}
\begin{alignat}{2}
    \text{MAB}^2_K(\mbx, \mby) &= \mbh + \text{fcc}(\text{relu}(\text{SetNorm}(\mbh)))\\
    \text{where } \mbh & = \mbx + \text{Attn}_K(\text{SetNorm}(\mbx),\text{SetNorm}(\mby), \mby).
\end{alignat}
Then, the ISAB++
block with  $D$ hidden units, $K$ heads and $M$ inducing points is defined as  
\begin{align}
    \text{ISAB++}_M(\mbx) &= \text{MAB}^2_K(\mbx, \mbh) \in \mathbb{R}^{S\times D}, \\
    \mbh &= \text{MAB}^1_K(\mbp, \mbx) \in \mathbb{R}^{M \times D}.
\end{align}

The reason why $\text{MAB}^1_K$ does not include normalization on the first input is because that inducing points $\mbp$ are learned.

%% file: ssections/flowrbc.tex
To complement our technical contributions, we open-source FlowRBC, a prototypical example of a clinically-available single cell blood dataset. In this type of dataset, permutation invariance holds biologically as blood cells move throughout the body.  
FlowRBC aims to answer an interesting physiological question: can we predict extrinsic properties from intrinsic ones? In practice, the task is to predict a patient's hematocrit levels from individual red blood cell (RBC) volume and hemoglobin measurements.
Hematocrit is the fraction of overall blood volume occupied by red blood cells and thus an aggregated measure of RBCs and other blood cell types. See more details in \cref{sec:hematocrit_description}.
FlowRBC represents an exciting real-world use case for prediction on sets largely overlooked by the machine learning community. It differs from other real-valued datasets  (e.g. Point Cloud) in that every absolute measurement carries biological information beyond its relative position with other points. This implies that translations might map to different physiological states. 
For this reason, careful architectural design
is required to preserve useful knowledge about the input.  

%% file: ssections/experiments.tex
To evaluate the effect of our proposed modifications, we consider
tasks with diverse inputs (point cloud, continuous, image) and outputs (regression, classification). 
We use four main datasets to study the individual components of our solution (Hematocrit, Point Cloud, Mnist Var and Normal Var)
and two (CelebA, Anemia) for validation of the models.
\begin{itemize}[leftmargin=*]
\item \textbf{Hematocrit Regression from Blood Cell Cytometry Data (Hematocrit a.k.a. Flow-RBC).}
\label{sec:hematocrit}
The dataset consists of measurements from 98240 train and 23104 test patients. 
We select the first visit for a given patient such that each patient only appears once in the dataset, and there is no patient overlap between train and test. We subsample for each distribution to 1,000 cells.

\item \textbf{Point Cloud Classification (Point Cloud).}
\label{sec:pc}
Following \cite{Zaheer2017DeepS, Lee2019SetTA}, we use the 
ModelNet40 dataset \citep{Wu20153DSA} (9840 train and 2468 test clouds),
randomly sample 1,000 points per set, and
standardize each object to have mean zero and unit variance along each coordinate axis. We report ablation results as cross entropy loss to facilitate the readability of the tables, i.e. lower is better.

\item \textbf{Variance Prediction, Image Data (MNIST Var).}
We implement empirical variance regression on MNIST digits as a proxy for real-world tasks with sets of images, e.g. prediction on blood smears or histopathology slides. We sample 10 images uniformly from the training set and use the empirical variance of the digits as a label. Test set and training set images are non-overlapping. Training set size is 50,000 sets, and test set size is 1,000 sets. We represent each image as a 1D vector.

\item \textbf{Empirical Variance Prediction, Real Data (Normal Var).}
\label{sec:nvm}
Each set is a collection of 1000 samples from a univariate normal distribution. Means are drawn uniformly in [-10, 10], and variances are drawn uniformly in [0, 10]. The target for each set is the empirical variance of the samples (regression task) in the set. Training set size is 10,000 sets, and test set size is 1,000 sets.
\item \textbf{Set anomaly detection, Image Data (CelebA).} Following \citet{Lee2019SetTA}, we generate sets of images from the CelebA dataset \citep{liu2015faceattributes} where nine images share two attributes in common while one does not. We learn an equivariant function whose output is a 10-dimensional vector that identifies the anomaly in the set.
We build a train and test datasets with 18000 sets, each of them containing 10 images (64$\times$64).
Train and test do not contain the same individuals. 

\item \textbf{Anemia detection, Blood Cell Cytometry
Data.} The dataset consists of 11136 train and 2432 test patients. Inputs are individual red blood cell measurements (volume and hemoglobin) and the outputs are a binary anemic vs. non-anemic diagnosis. A patient was considered anemic if they had a diagnosis for anemia of any type within 3 days of their blood measurements. We sample 1,000 cells for each input distribution.
\end{itemize}

Unless otherwise specified, results are reported in Mean Squared Error (MSE) for regression experiments and in cross entropy loss (CE) for point cloud classification, averaged over three seeds. We fix all hyperparameters, including epochs, and use the model at the end of training for evaluation. We notice no signs of overfitting from the loss curves. 
For further experimental details, see \Cref{sec:experimental_configuration}.

%% file: ssections/results.tex
\begin{table*}[]
    \centering  
        \caption{Equivariant residual connections perform better than aggregated residual connections in both Deep Sets and Set Transformer.  Max aggregation for Set Transformer led to exploding gradient so we do not report result. }
          \resizebox{0.9\textwidth}{!}{ \begin{tabular}{llcccc}
      \toprule
 Path & Residual type  & Hematocrit (MSE) & Point Cloud (CE) & Mnist Var (MSE) & Normal Var (MSE) \\
         \hline
 \rowcolor{lightgray} Deep Sets  & equivariant 
          &\textbf{19.2118} $\pm $ \textbf{0.0762} & \textbf{0.7096} $\pm$ \textbf{0.0049} &\textbf{ 0.3441} $\pm$ \textbf{0.0036}
& \textbf{0.0198} $\pm$ \textbf{0.0041} \\
& mean & 
  19.3462 $\pm$ 0.0260 & 0.8585 $\pm$ 0.0253 & 1.2808 $\pm$ 0.0101 & 
0.8811 $\pm$ 0.1824 \\
 &  max & 
19.8171 $\pm$  0.0266 &0.8758 $\pm$ 0.0196 &1.3798$ \pm$ 0.0162 & 
0.8964 $\pm$ 0.1376 \\
\hline 
  \rowcolor{lightgray}Set Transformer  &equivariant 
        & \textbf{18.6883} $\pm$ \textbf{0.0238}  & \textbf{0.6280} $\pm $\textbf{0.0098} & \textbf{0.7921} $\pm$ \textbf{0.0006} & 
        \textbf{0.0030} $\pm$ \textbf{0.0000} \\
      &  mean
         & 19.6945 $\pm$ 0.1067 & 0.8111 $\pm$ 0.0453 & 1.6273 $\pm$ 0.0335 & 
        0.0147 $\pm$ 0.0028 \\
        \bottomrule
    \end{tabular}}
    \label{tab:type_res}
\end{table*}

\begin{table*}[t]
    \centering
    \caption{While Deep Sets and Set Transformer exhibit notable failures when deep (underlined), \glsfirst{ds++} and \glsfirst{st++} do not. The latter 
    also
    achieve new levels of performance on a several tasks.}
  \label{tab:main_results}
  \centering
  \resizebox{\textwidth}{!}{
  \begin{tabular}{ll cc|ccc}
    \toprule
         Model & No. Layers & Hematocrit (MSE) & MNIST Var (MSE)  & Point Cloud (accuracy) & CelebA (accuracy) & Anemia (accuracy)\\
         \midrule
       DeepSets  &3  & \textbf{19.1257 $\pm$ 0.0361} &  0.4520 $\pm $0.0111 & 0.7755 $\pm$ 0.0051  &0.3808 $\pm$ 0.0016 & 0.5282 $\pm$ 0.0018 \\
        &25 & 20.2002 $\pm$ 0.0689 & 1.3492 $\pm$ 0.2801 & \underline{0.3498 $\pm$ 0.0340} &  \underline{0.1005 $\pm$ 0.0000} & \underline{0.4856 $\pm$ 0.0000} \\
        &50 & \underline{25.8791$\pm$ 0.0014}  &  \underline{5.5545 $\pm$ 0.0014} & \underline{0.0409 $\pm$ 0.0000} &\underline{0.1005 $\pm$ 0.0000}  & \underline{0.4856 $\pm$ 0.0000}\\
            Deep Sets++ &3 &19.5882 $\pm$ 0.0555 &  0.5895 $\pm$ 0.0114 &0.7865 $\pm$ 0.0093  &  0.5730 $\pm$ 0.0016 & 0.5256 $\pm$ 0.0019 \\
         &25 & \textbf{19.1384 $\pm$ 0.1019} &  0.3914 $\pm$ 0.0100 &\textbf{0.8030 $\pm$ 0.0034} &\textbf{0.6021 $\pm$ 0.0072} & 0.5341 $\pm$ 0.0118\\
         \rowcolor{lightgray}
         &50 & \textbf{19.2118 $\pm $ 0.0762} &\textbf{0.3441 $\pm$ 0.0036} & \textbf{0.8029 $\pm$ 0.0005} & \textbf{0.5763 $\pm$ 0.0134} & \textbf{0.5561 $\pm$ 0.0202}\\
        \midrule
        Set Transformer &2  & 18.8750 $\pm$ 0.0058 &  0.6151 $\pm$ 0.0072 & 0.7774 $\pm$ 0.0076 &0.1292 $\pm $ 0.0012 & \textbf{0.5938 $\pm$ 0.0075}\\
        \rowcolor{lightgray}
        &8  & 18.9095 $\pm$ 0.0271 &  \textbf{0.3271 $\pm$ 0.0068} & 0.7848 $\pm$ 0.0061 &0.4299 $\pm$ 0.1001 & \textbf{0.5943 $\pm$ 0.0036}\\
        &16  & \textbf{18.7436 $\pm$ 0.0148} & \underline{6.2663 $\pm$ 0.0036} &  \underline{0.7134 $\pm$ 0.0030} &0.4570 $\pm $ 0.0540 & 0.5853 $\pm$ 0.0049\\
        Set Transformer++ &2 & 18.9223 $\pm$ 0.0273 & 1.1525 $\pm$ 0.0158 &0.8146 $\pm$ 0.0023 &  0.6533 $\pm$ 0.0012 & 0.5770 $\pm$ 0.0223\\
         \rowcolor{lightgray}
         &8  & 18.8984 $\pm$ 0.0703 & 0.9437 $\pm$ 0.0137 &\textbf{0.8247 $\pm$ 0.0020} &\textbf{0.6621 $\pm$ 0.0021} & 0.5680 $\pm$ 0.0110\\
         \rowcolor{lightgray}
         &16  &  \textbf{18.7008  $\pm$ 0.0183} &0.8023 $\pm$ 0.0038 & \textbf{0.8258 $\pm $ 0.0046} & 0.6587 $\pm$ 0.0001 & 0.5544 $\pm$ 0.0113\\
        \bottomrule
    \end{tabular}}
\end{table*}

\paragraph{Clean path residuals have better performance than non-clean path ones.} 
\Cref{tab:clean_nonclean} confirms 
that clean path pipelines generally yield the best performance across set tasks both for Deep Sets and Set Transformer, independently of normalization choice. The primary exception to this trend is Deep Sets on Point Cloud, which can be explained by a Point Cloud-specific phenomenon where the repeated addition of positive values in the architecture improves performance (see \Cref{subseq:pointcloud} for empirical analysis). 
Non-clean path Set Transformer has both the worst and best results on Mnist Var among Set Transformer variants, evidence of its unpredictable behavior at high depths. In contrast, ST++ results are more stable, and \Cref{tab:main_results} illustrates that ST++ consistently improves on Mnist Var as depth increases.

The clean path principle has previously been shown in other applications to improve performance and yield more stable training \citep{He2016IdentityMI, wang2019deeptransformers}.
Its benefit for both Deep Sets and Set Transformer provides further proof of the effectiveness of this principle.

\paragraph{Equivariant residual connections are the best choice for set-based skip connections.}
ERCs
generalize residual connections to permutation-equivariant architectures. 
For further validation of their usefulness, we empirically compare them with 
another type of residual connection: an \emph{aggregated residual connection} (ARC) which sums an aggregated function of the elements (e.g. sum, mean, max) from the previous layer. 
\Cref{sec:arc} provides a more detailed discussion. 
Results in \Cref{tab:type_res}  show that clean-path ERCs remain the most suitable choice.

\paragraph{Set norm performs better than other norms.}
\Cref{tab:ds_norm} shows that Deep Sets benefits from the addition of set norm  when no residual connections are involved. Hematocrit and Normal Var performances are the same across normalizations, but this is due to a vanishing gradient that cannot be overcome by the presence of normalization layers alone.

We further analyzed normalizations in the presence of residual connections in \Cref{tab:clean_nonclean}.
Here, we also consider 
the normalization layer used in the PointNet and PointNet++ architectures for point cloud classification \citep{Qi2017PointNetDL, Qi2017PointNet++}, implemented as batch norm on a transposed tensor.
We call this norm \emph{feature norm}, 
which is an example of a normalization that occurs over the batch rather than on a per-set basis ($\mathcal{S} = \{D\}, \mathcal{T} = \{D\})$.

Clean path residuals with set norm generally perform best. The pattern is particularly evident for Normal Var,
where
clean path is significantly better than non-clean path and the addition of set norm further improves the performance.

We additionally observe in \Cref{tab:clean_nonclean} that results for layer norm improve with the addition of clean-path residual connections relative to earlier results in \Cref{tab:layer_norm} and \Cref{tab:ds_norm}.
We hypothesize that skip connections help alleviate information loss from normalization by passing forward values before normalization. 
For instance, given two elements $\mbx_l$ and $\mbx'_l$ that will be mapped to the same output $\hat{\mbx}$ by layer norm, adding a residual connection enables the samples to have distinct outputs $\mbx_{l+1} = \mbx_l + \hat{\mbx} $ and $\mbx'_{l+1} = \mbx'_l + \hat{\mbx}$.

\paragraph{Deep Sets++ and Set Transformer++ outperform existing architectures.}
We validate our proposed models \gls{ds++} and \gls{st++} on real-world datasets (\Cref{tab:main_results}). Deep Sets (DS) and Set Transformer (ST) show failures (underlined entries) as depth increases. On the contrary,  \gls{ds++} and \gls{st++} tend to outperform their original and shallow counterparts at high depths (rows highlighted in gray have the highest number of best results).
Deep Sets++ and Set Transformer++ particularly improve performance on point cloud classification and CelebA set anomaly detection. 
We show in \Cref{sec:add_results} that, on an official point cloud benchmark repository \citep{Goyal2021RevisitingPC},  \gls{ds++} and \gls{st++} without any modifications outperform versions of Deep Sets and Set Transformer tailored for point cloud classification.
On Hematocrit, both deep modified models surpass the clinical baseline (25.85 MSE) while 
the original Deep Sets at 50 layers does not
(more details are provided in \Cref{sec:hematocrit_description}). 

\Cref{tab:main_results} highlights that
 \gls{ds++} and \gls{st++} generally improve over DS and ST overall 
without 
notable failures as depth increases. 
Due to their reliability and ease of use, \gls{ds++} and \gls{st++} are practical choices for practitioners who wish to avoid extensive model search or task-specific engineering when approaching a new task, particularly one involving sets of measurements or images. We expect this benefit to be increasingly relevant in healthcare or biomedical settings, as new datasets of single cell measurements and cell slides continue to be generated, and new tasks and research questions continue to be posed.  

Lastly, while ST and \gls{st++} performance are better than \gls{ds++}, it is worth noticing that the former models have approximately 3 times more parameters and take more time and memory to run. As an example, on point cloud classification, \gls{st++} took $\approx 2$ times longer to train than \gls{ds++} for the same number of steps on a NVIDIA Titan RTX. 

%% file: ssections/related.tex
Previous efforts to design residual connections \citep{he2016resenet, Veit2016ResidualNB, Yao2020PyHessianNN} or normalization layers \citep{Ioffe2015BatchNA, Ba2016LayerN, Santurkar2018HowDB,Ghorbani2019AnII, Luo2019TowardsUR,Xiong2020OnLN, Cai2021GraphNormAP} have often been motivated by particular applications. Our work is motivated by applications of predictions on sets.

The effects of non-clean or clean path residual connections have been studied in various settings. \citet{He2016IdentityMI} showed that adding a learned scalar weight to the residual connection, i.e. $\mbx_{\ell+1} = \lambda_\ell \mbx_\ell + \mathcal{F}(\mbx_\ell)$, can result in vanishing or exploding gradients
if the $\lambda$ scalars are consistently large (e.g. $> 1$) or small ($< 1$).
\citet{wang2019deeptransformers} 
while
see that for deep transformers, only the clean path variant converges during training.
\citet{Xiong2020OnLN} show that Post-LN transformers (non-clean path) require careful learning rate scheduling unlike their Pre-LN (clean path) counterparts.
Our analysis provides further evidence of the benefit of clean-path residuals.
While our clean and non-clean path DS architectures mirror those of the clean and non-clean path ResNet architectures \citep{He2016IdentityMI, he2016resenet}, the 
non-clean path Set Transformer differs from non-clean path Post-LN Transformer in that the former also has a linear projection on the residual path.

Many normalization layers have been designed for specific purposes.
For instance, batch norm \citep{De2020BatchNB}, layer norm \citep{Ba2016LayerN}, instance norm \citep{Ulyanov2016InstanceNT} and graph norm \citep{Cai2021GraphNormAP} were designed for image, text, stylization, and graphs respectively.
In this work, we propose set norm with set inputs in mind, particularly sets of real-valued inputs.
The idea in set norm to address different samples being mapped to the same outputs from layer norm
is reminiscent of the goal to avoid oversmoothing motivating pair norm
\cite{zhao2019pairnorm}, developed for graph neural networks.

Our work offers parallels with work on graph convolutional networks (GCNs). For instance, previous works in the GCN literature have designed architectures that behave well when deep
and leverage residual connections
\citep{Guohao2019deepGCN, chen2020simple}. 
However, while GCNs and set-based architectures share a lot of common principles, the former relies on external information about the graph structure which is not present in the latter.

%% file: ssections/conclusions.tex
We illustrate limitations of Deep Sets and Set Transformer when deep and develop \glsfirst{ds++} and \glsfirst{st++}
to overcome these limitations. We introduce \glsfirst{ipn} to address the unwanted invariance of layer norm for real-valued sets, and we employ clean-path equivariant residual connections to enable identity mappings and help address gradient issues.
\gls{ds++} and \gls{st++} are general-purpose architectures and the first permutation invariant architectures of their depth that show good performance on a variety of tasks. 
We also introduce Flow-RBC, a new open-source dataset which provides a real-world application of permutation invariant prediction in clinical science. 
We believe our new models and dataset have the potential to motivate future work and applications of prediction on sets.

%% file: appendix/hematocrit.tex
\clearpage
\section{Flow-RBC}\label{sec:hematocrit_description}
The analysis of and prediction from single-cell data is an area of rapid growth \citep{Lhnemann2020ElevenGC}. Even so, Flow-RBC constitutes a dataset unique for its kind, consisting of more than 100,000 measurements taken on different patients paired with a clinical label. Even established projects like the Human Cell Atlas \citep{regev2017science} or Flow Repository\footnote{https://flowrepository.org} do not include single-cell datasets of this size. For instance, to our knowledge, the second largest open-source dataset of single-cell blood samples contains data from 2,000 individuals and does not include external clinical outcomes for all patients to be used as a target.

\begin{figure}[h]
    \centering
    \includegraphics[width=0.5\textwidth]{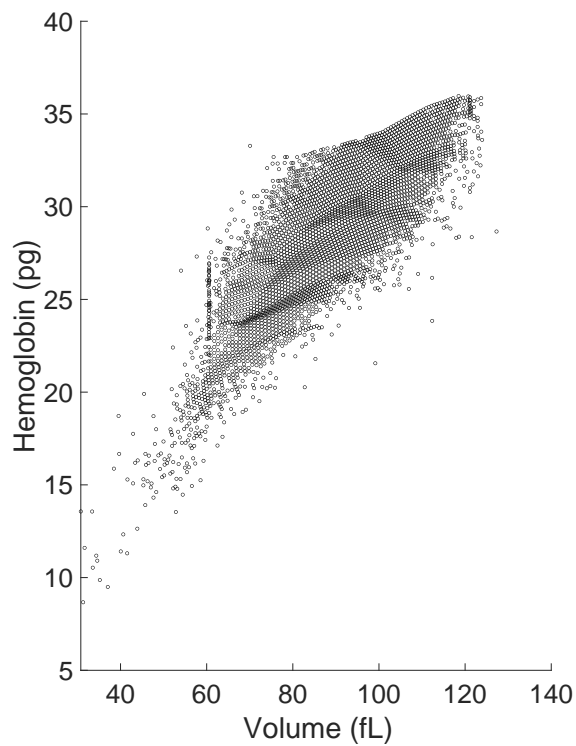}
    \caption{Example of RBC distribution given in input for the prediction of hematocrit level. }
    \label{fig:rbcdistr}
\end{figure}

Flow-RBC consists of 98,240 train and 23,104 test examples. Each input set is a red blood cell (RBC) distribution of 1,000 cells. Each cell consists of a volume and hemoglobin content measurement (see \Cref{fig:rbcdistr} for a visual representation). The regression task consists of predicting the corresponding hematocrit level measured on the same blood sample. Blood consists of different components: red blood cells, white blood cells, platelets and plasma. The hematocrit level measures the percentage of volume taken up by red blood cells in a blood sample.

Since we only have information about the volume and hemoglobin of individual RBCs and no information about other blood cells,
this task aims to answer an interesting clinical question: is there information present in individual RBC volume and hemoglobin measurements about the overall volume of RBCs in the blood? As this question has not been definitively answered in the literature, there is no known expected performance achievable; instead, increases in performance are an exciting scientific signal, suggesting a stronger relationship between single cell RBC and aggregate population properties of the human blood than previously known. 

The existing scientific literature notes that in the presence of diseases like anemia, there exists a 
negative correlation between hematocrit and the red cell distribution width (RDW), also known as the coefficient of variation of the volume i.e. SD(Volume) / Mean(Volume) $\times$ 100 \citep[Chapter~9]{mcpherson2021henry}. 
To represent the current state of medical knowledge on this topic, we use as a baseline
a linear regression model with RDW as covariate. 
Additionally, we build a regression model on hand-crafted distribution statistics (up to the fourth moment on both marginal distributions as well as .1, .25, .5, .75, .9 quantiles). This model improves over simple prediction with RDW, further confirming the hypothesis that more information lies in the single-cell measurements of RBCs. \Gls{st++} further improves performance, resulting in an MSE reduction of 28\% over the RDW model. See \Cref{tab:baselinehematocrit} for results.

\begin{table}[h]
    \centering
      \caption{Baseline regression performances for the prediction of hematocrit from RBC distributions. Our proposed Set Transformer++ currently has the best performance on this task. }
    \label{tab:baselinehematocrit}
    \begin{tabular}{l|c}
    \toprule
         & MSE  \\
         \midrule
         RDW & 25.85\\
         Moments & 22.31 \\
         Set Transformer++ & 18.69\\
         \bottomrule
    \end{tabular}
  
\end{table}

\paragraph{Procedure to Obtain RBC distribution measurements}
All  Flow-RBC data is collected retrospectively at 
Massachusetts General Hospital
under an existing IRB-approved research protocol and is available at \href{https://cims.nyu.edu/~lhz209/flowrbc}{this link}. 
Each RBC distribution consists of volume and hemoglobin mass measurements collected using the Advia 2120 \citep{harris2005advia}, a flow-cytometry based system that measures thousands of cells. The volume and hemoglobin information are retrieved through Mie (or Lorenz-Mie) theory equations for the analysis of light scattering from a homogeneous spherical particle \citep{tycko1985flow}. An example of one input distribution is provided in \Cref{fig:rbcdistr}. The Advia machine returns an average of 55,000 cells. For this dataset, we downsampled each distribution to 1,000 cells, a number high enough to maintain sample estimates of ``population'' (i.e. all 55,000 cells) statistics with minimal variance while imposing reasonable memory requirements on consumer gpus.
Each distribution is normalized and re-scaled by the training set mean and standard deviation.

%% file: appendix/layer_norm.tex
\section{Layer Norm Analyses}

\subsection{Gradient Analysis for Set Transformer with Layer Norm}\label{sec:st_layernorm_ga}
The addition of Layer norm to \Cref{eq:transformer} does not preclude the possibility of
exploding or vanishing gradients. 
Let $\text{Attn}_K$ be multihead attention with $K$ heads and a scaled softmax, \textit{i.e.} $\text{softmax}(\cdot / \sqrt{D})$, and let LN be layer norm. We consider the following definition of a MAB module, with layer norm placement that matches what was described in the original paper \citep{Lee2019SetTA}:
\begin{align}
    \text{MAB}_K(\mbx, \mby) &= \text{LN}(f(\mbx, \mby) + \text{relu}(f(\mbx, \mby) W + \mathbf{b})), \label{eq:ff2}\\
    f(\mbx, \mby) &= \text{LN}(\mbx W_Q + \text{Attn}_K(\mbx, \mby, \mby)).
\end{align}
The inducing point set attention block (ISAB) is then
\begin{align}
    ISAB_M(\mbx) &= \text{MAB}_K(\mbx, \mbh) \in \mathbb{R}^{S\times D} \\
    \text{where } \mbh &= \text{MAB}_K(\mbp, \mbx) \in \mathbb{R}^{M \times D}.
\end{align}

Consider a single ISAB module. We let $\mbz_1$ denote the output of the previous block, $\mbz_2$ denote the output after the first MAB module (i.e. $\mbh$ in \Cref{eq:first_mab}), and $\mbz_3$ denote the output of the second MAB module, or the overall output of the ISAB module. Then,
\begin{align}
    f_1 &= f(\mathbf{p}, \mbz_1) = \text{LN}(I W_{1Q} + \text{Attn}_K(\mathbf{p}, \mbz_1, \mbz_1)) \\
    \mbz_2 &= \text{LN}(f_1 + \text{relu}(f_1 W_1 + \mathbf{b}_1)) \\
    f_2 &= f(\mbz_1, \mbz_2) =\mbz_1 W_{2Q} + \text{Attn}_K(\mbz_1, \mbz_2, \mbz_2)) \\
    f_3 &=  \text{LN}(f_2)\\
    f_4 &=  f_3 + \text{relu}(f_3 W_2 + \mathbf{b}_2)\\
    \mbz_3 &=\text{LN}(f_4).
\end{align}
The gradient of a single ISAB block output $\mbz_3$ with respect to its input $\mbz_1$ can be represented as $\frac{\partial{\mbz_3}}{\partial{\mbz_1}} = \frac{\partial{\mbz_3}}{\partial{f_4}} 
\frac{\partial{f_4}}{\partial{f_3}}
\frac{\partial{f_3}}{\partial{f_2}} 
\frac{\partial{f_2}}{\partial{\mbz_1}}$, or
\begin{align*}
    \frac{\partial{\text{LN}(f_4)}}{\partial{f_4}}\Big(I + \frac{\partial{\text{relu}(f_2 W_2 + \mathbf{b}_2)}}{\partial{(f_2 W_2 + \mathbf{b}_2)}}W_2\Big)\\
    \frac{\partial{\text{LN}(f_2)}}{\partial{f_2}}\Big(
    W_{2Q}+ \frac{\partial{\text{Attn}_K(\mbz_1, \mbz_2, \mbz_2)}}{\partial{\mbz_1}}\Big).
\end{align*}
The gradient expression is analogous to the one in \Cref{sec:settransformer_exploding}, with the exception of additional $\frac{\partial{\text{LN}(f_4)}}{\partial{f_4}}$ and $\frac{\partial{\text{LN}(f_2)}}{\partial{f_2}}$ per ISAB block. With many ISAB blocks, it is still possible for a product of the weights $W_{2Q}$ to accumulate.

\subsection{Visualizing Layer Norm Example in 2D}
\label{sec:2d_ln}
\begin{figure}[H]
    \centering
    \includegraphics[width=.9\linewidth]{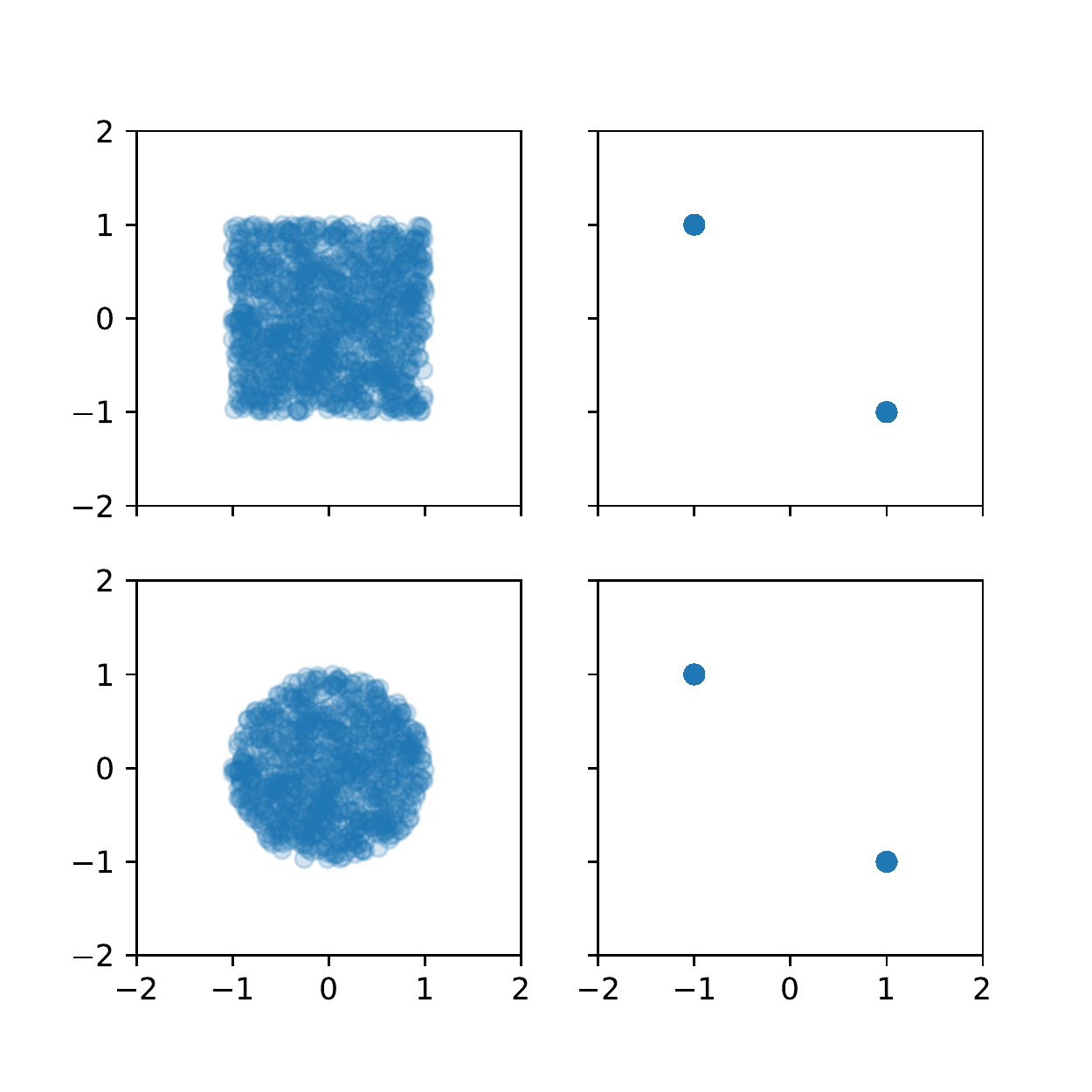}
    \caption{Layer norm performs per-sample standardization, which in 2D point cloud classification can result in shapes (left) whose 2D activations (right) are indistinguishable from each other.}
    \label{fig:layer_norm_shapes_2d}
\end{figure}

In \Cref{sec:layernorm_issues}, we discussed how layer norm removes two degrees of freedom from each sample in a set, which can make certain prediction difficult or impossible. In particular, we discussed a simple toy example in 2D, that of classifying shapes based on 2D point clouds. We utilize hidden layers of size 2, which means the resulting activations can be visualized. In this setup, different shapes yield the same resulting activations as long as their points are equally distributed above and below the $y=x$ line. See \Cref{fig:layer_norm_shapes_2d}.

%% file: appendix/norm_proofs.tex
\section{Normalization proofs}
\label{sec:norm_proofs}

\textbf{Proposition 1.} \textit{Let $\mathcal{F}$ be the family of transformation functions which can be expressed via \Cref{eq:transform}.
Then, for $f \in \mathcal{F}$, $\mathcal{T} = \{D\}$ and $\mathcal{T}=\{\}$ are the only settings satisfying the following properties: 
\begin{enumerate}
    \item $f_\mathcal{T}(\pi_i \mba) = \pi_i f_\mathcal{T}(\mba)$ where $\pi_i$ is a permutation function that operates on elements in the set; 
    \item $f_\mathcal{T}(\pi_n \mba) = \pi_n f_\mathcal{T}(\mba)$ where $\pi_n$ is a permutation function that operates on sets.
\end{enumerate}}

\begin{proof}
For transformation tensors in $\mathbb{R}^{N \times M \times D}$, the parameters can be distinct over the batch $(N \in \mathcal{T})$, over the elements ($M \in \mathcal{T}$), over the features $(D \in \mathcal{T})$, or any combination of the three. We show that $N \in \mathcal{T}$ and $M \in \mathcal{T}$ are unsuitable, leaving only $D \in\mathcal{T}$.

Having distinct parameters over the samples breaks permutation equivariance, making $M \in \mathcal{T}$ an untenable option. Let $f: \mathbb{R}^{M \times D} \rightarrow \mathbb{R}^{M \times D}$ be the transformation function, and $\overrightarrow{\gamma}_{\{M\}}, \overrightarrow{\beta}_{\{M\}}$ represent tensors in $\mathbb{R}^{N\times M \times D}$ where the values along dimension $M$ can be unique, while the values along $N, D$ are repeated. We denote an indexing into the batch dimension as $\overrightarrow{\gamma}_{\{M\},n}, \overrightarrow{\beta}_{\{M\},n}$. Then, $f$ breaks permutation equivariance:

\begin{align}
    f(\pi_i \mba) &= \pi_i \mba \odot \overrightarrow{\gamma}_{\{M\},n} + \overrightarrow{\beta}_{\{M\},n} \\
    &\neq \pi_i ( \mba \odot \overrightarrow{\gamma}_{\{M\},n} + \overrightarrow{\beta}_{\{M\},n}) \\
    &= \pi_i f(\mba).
\end{align}

Having distinct parameters over the batch means that the position of a set in the batch changes its ordering,  making $N \in \mathcal{T}$ an untenable option. Let $\overrightarrow{\gamma}_{\{N\}}, \overrightarrow{\beta}_{\{N\}}$ represent tensors which can differ over the batch, e.g. $\overrightarrow{\gamma}_{\{N\}, n} \neq \overrightarrow{\gamma}_{\{N\}, n'}, n \neq n'$. Then, the prediction function $f_n$ for batch index $n$ will yield a different output than the prediction function $f_{n'}$ for batch index $n'$:
\begin{align}
    f_b(\mba) &= \mba \odot \overrightarrow{\gamma}_{\{N\},n} + \overrightarrow{\beta}_{\{N\},n} \\
    &\neq \mba \odot \overrightarrow{\gamma}_{\{N\},n'} + \overrightarrow{\beta}_{\{N\},n'} \\
    &= f_{n'}(\mba).
\end{align}

As neither $M$ nor $N$ can be in $\mathcal{T}$, the remaining options are $\mathcal{T} = \{D\}$ or $\mathcal{T}= \{\}$, i.e. $\overrightarrow{\gamma}, \overrightarrow{\beta}$ each repeat a single value across the tensor. Note that $\mathcal{T} = \{\}$ is strictly contained in $\mathcal{T} = \{D\}$: if the per feature parameters are set to be equal in the $\mathcal{T} = \{D\}$ setting, the result is equivalent to $\mathcal{T} = \{\}$. Therefore, $\mathcal{T} = \{D\}$ sufficiently describes the only suitable setting of parameters for transformation.

\end{proof}

\begin{proposition}{Set norm is permutation equivariant.}
\end{proposition}

\begin{proof}
Let $\mu, \sigma \in \mathbb{R}$ be the elements mean and variance over all features in the set, $\overrightarrow{\gamma}, \overrightarrow{\beta} \in \mathbb{R}^{M \times D}$ refer to the appropriate repetition of per-feature parameters in the $M$ dimension. Then,
\begin{align}
\text{SN}(\pi \mbx)
&= \frac{\pi \mbx - \mu}{\sigma} * \overrightarrow{\gamma} + \overrightarrow{\beta} ] \\    
&=\pi \Big[\frac{\mbx - \mu}{\sigma} * \overrightarrow{\gamma} + \overrightarrow{\beta}\Big] \label{eq:move_out_pi} \\    
&= \pi \text{SN}(\mbx).
\end{align}

\Cref{eq:move_out_pi} follows from the fact that $\mu, \sigma$ are scalars and $\overrightarrow{\gamma}, \overrightarrow{\beta}$ are equivalent for every sample in the set.
\end{proof}

%% file: appendix/experiments.tex
\section{Experimental configuration} \label{sec:experimental_configuration}
Across experiments and models we purposefully keep hyperparameters consistent to illustrate the easy-to-use nature of our proposed models. All experiments and models are implemented in PyTorch. The code is available at \href{https://github.com/rajesh-lab/deep_permutation_invariant}{https://github.com/rajesh-lab/deep\_permutation\_invariant}.

\subsection{Experimental Setup}
Hematocrit, Point Cloud and Normal Var use a fixed sample size of 1000. MNIST Var and CelebA use a sample size of 10 due to the high-dimensionality of the images in input. The only architectural difference across these experiments is the choice of permutation-invariant aggregation for the Deep Sets architecture: we use sum aggregation for all experiments except Point Cloud, where we use max aggregation, following 
\citep{Zaheer2017DeepS}. We additionally use a featurizer of convolutional layers for the architectures on CelebA given the larger image sizes in this task (see \Cref{sec:celeba} section for details).

All models are trained with a batch size of 64 for 50 epochs, except for Hematocrit where we train for 30 given the much larger size of the training dataset (i.e. 90k vs. $\leq$ 10k). All results are reported as test MSE (or cross entropy for point cloud) at the last epoch. We did not use early stopping and simply took the model at the end of training. There was no sign of overfitting. Results are reported setting seeds 0, 1, and 2 for initialization weights.   
We use the Adam optimizer with learning rate 1e-4 throughout.

\subsection{Convolutional blocks for set anomaly}
\label{sec:celeba}
For our set anomaly task on CelebA, similarly to \citet{Zaheer2017DeepS}, we add at the beginning of all the considered architectures 9 convolutional layers with $3 \times 3$ filters. Specifically, we start with 2D convolutional layers with 32, 32, 64  feature-maps followed by max pooling; we follow those with 2D convolutional layers with 64, 64, 128 feature maps followed by another max pooling; and  weend with 128, 128, 256 2D convolutional layers followed by a max-pooling layer with size 5. The output of the featurizer (and input to the rest of the permutation invariant model) is 255 features. The architecture is otherwise the same as those used on all other tasks considered in this work.

%% file: appendix/additional_results.tex
\section{Additional results}
\label{sec:add_results}

\subsection{Understanding ResNet vs. He Pipeline for Deep Sets on Point Cloud.} \label{subseq:pointcloud}
\begin{figure*}
    \centering
    \includegraphics[width=0.45\textwidth]{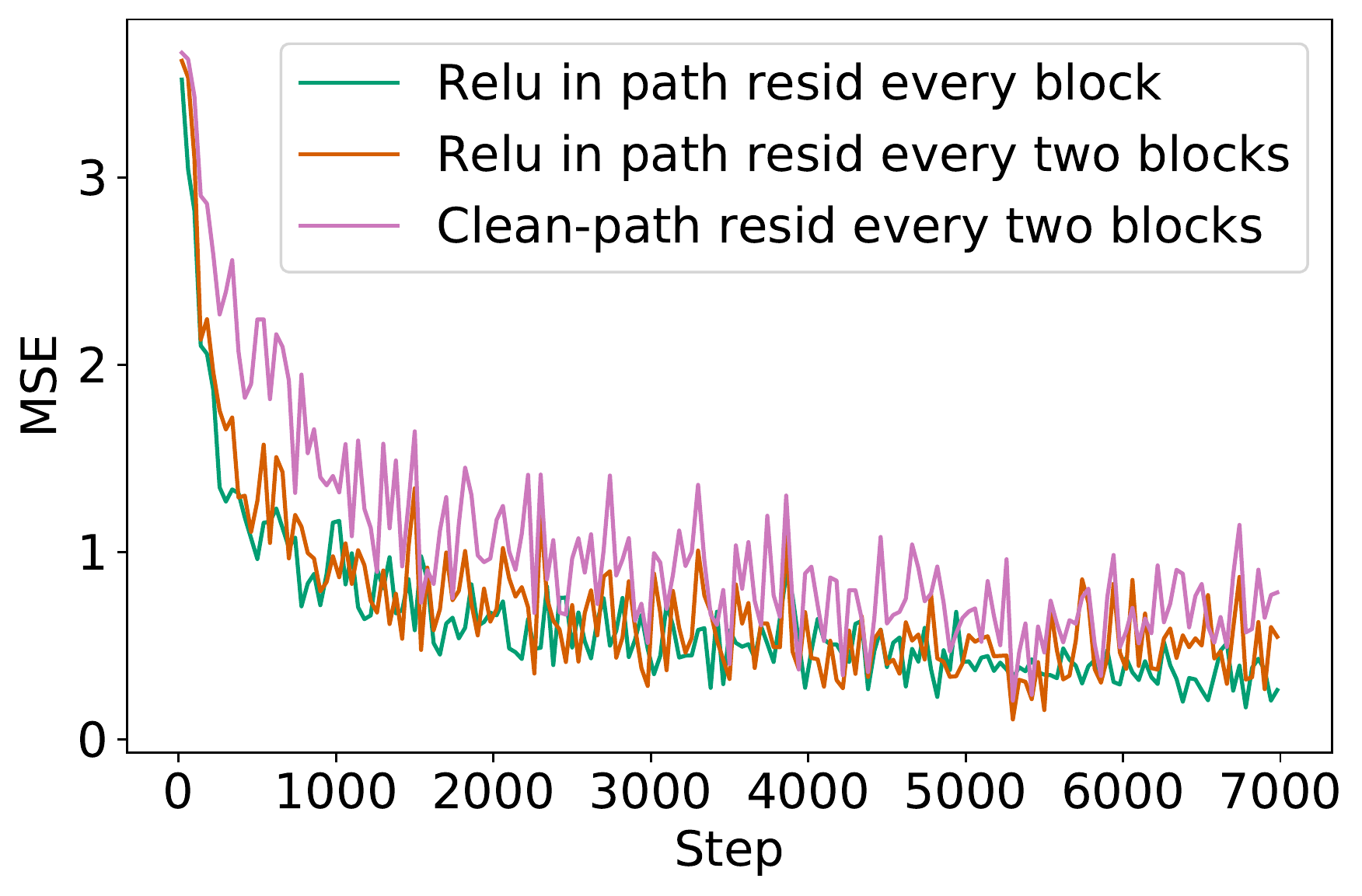}
    \includegraphics[width=0.47\textwidth]{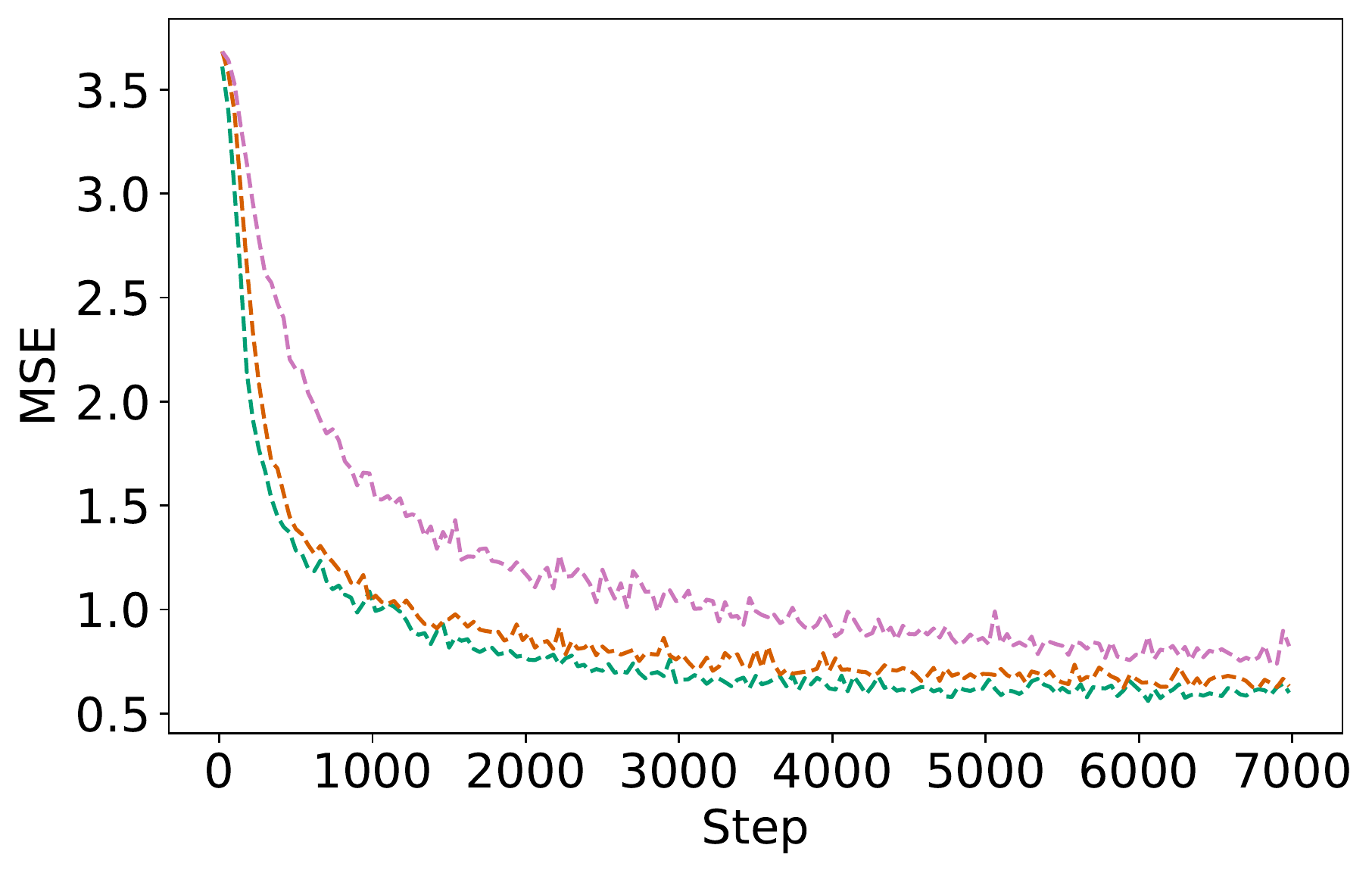}
    \caption{Loss curves for train (left) and test (right) comparing the residual pipelines ResNet (orange), He (magenta) and FreqAdd (green). Adding a positive number more frequently (green $>$ orange $>$ magenta) results in better performance for Point Cloud.}
    \label{fig:more_res_pointcloud}
\end{figure*}

We explore why Deep Sets with the non-clean ResNet residual pipeline performs better on Point Cloud than Deep Sets with the clean He residual pipeline. Specifically, to test whether the difference is due to the ReLU activation in between connections, we design another residual pipeline where the connections (i.e. additions) are more frequent and also separated by a ReLU nonlinearity. We call this pipeline FreqAdd. This new architecture is shown in \Cref{tab:more_frequent_res} and comparison of loss curves is in \Cref{fig:more_res_pointcloud}
where we can observe that the architecture with more residual connection FreqAdd has even better performances than the non-clean pipeline. We speculate that this might be due to peculiarities of Point Cloud which benefit from continual addition positive values. Indeed, in the original Deep Sets paper \citep{Zaheer2017DeepS}, the authors add a ReLU to the end of the encoder for the architecture tailored to point cloud classification, and such a nonlinearity is noticeably missing from the model used for any other task.

\begin{table}
 \caption{Detailed DeepSets more residuals architecture.}
 \label{tab:more_frequent_res}
 \centering
    \resizebox{0.48\textwidth}{!}{\begin{tabular}{cc|c|c}
    \hline
      \multicolumn{2}{c}{Encoder} & Aggregation & Decoder\\
    \hline
        Residual block $\times$ 51 & & &\\
        \hline
         FC(128) & FC(128) & Sum/Max & FC(128)\\ 
         SetNorm(128) & & & ReLU \\
          Addition  &  & & FC(128)\\
         ReLU& & &ReLU\\
         & & &FC(128)\\
         & && ReLU\\
         & && FC(no\_outputs)\\
        \hline
    \end{tabular}}
\end{table}

\subsection{Comparing Point Cloud classification with Task-Specific Models}
\label{sec:pc_tailor}
Here, we compare the performances of \gls{ds++} and \gls{st++} unmodified with those of models built specifically for point cloud classification. 
For a fair comparison, we use the experimental
setup and the code provided in SimpleView  \citep{goyal2021SimpleView}. In practice, we use their DGCNN-smooth protocol and record the test accuracy at 160 epochs. The sample size for this experiment is the default in the SimpleView repository, 1024. We compared Deep Sets++, Set Transformer++, PointNet++ \citep{Qi2017PointNet++}, and SimpleView \citep{Goyal2021RevisitingPC}, as well as the models proposed in the original Deep Sets and Set Transformer papers tailored to point cloud classification, which differ than from the baseline architectures used in our main results.  We describe these tailored Deep Sets and Set Transformer models in \Cref{tab:deepsets_pointcloud} and \Cref{tab:settransformer_pointcloud}.

\begin{figure}
    \centering
    \includegraphics[width=0.5\textwidth]{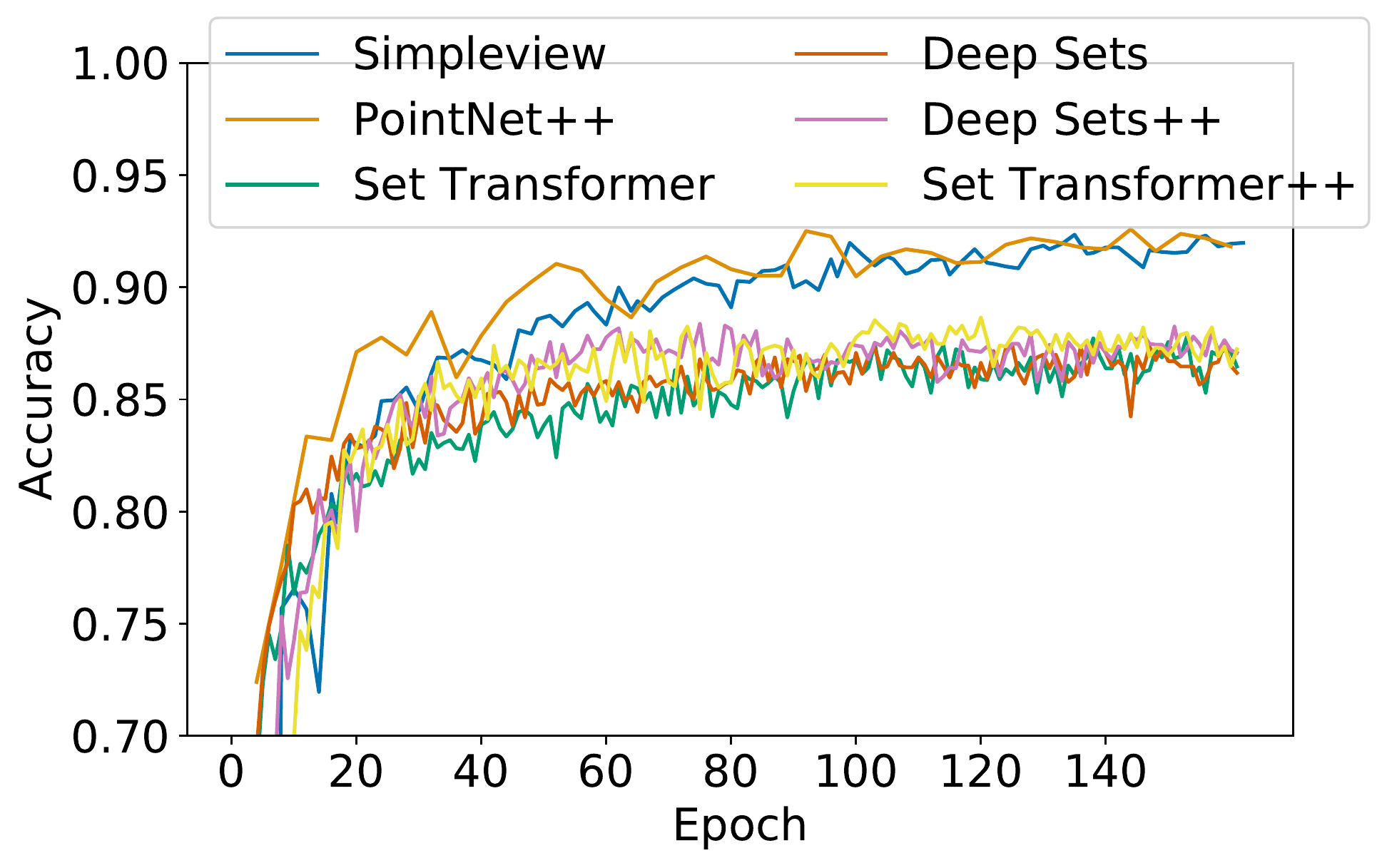}
    \caption{Test accuracy curves of different architectures on Point Cloud classification, as implemented in the SimpleView codebase. Unmodified DS++ and ST++ outperform DS and ST tailored to the task.}
    \label{fig:simpleview_acc}
\end{figure}

Results are reported in \Cref{tab:pointcloud_acc} and \Cref{fig:simpleview_acc}. Deep Sets++ and Set Transformer++ without any modifications both achieve a higher test accuracy than the Deep Sets and Set Transformer models tailor designed for the task. PointNet++ and SimpleView perform best, but both architectures are designed specifically for point cloud classification rather than tasks on sets in general. Concretely, PointNet++ hierarchically assigns each point to centroids using Euclidean distance which is not an informative metric for high-dimensional inputs, e.g. sets of images.
SimpleView is a non-permutation invariant architecture that represents each point cloud by 2D projections at various angles; such a procedure is ill-suited for sets where samples do not represent points in space.

\begin{table}[]
\centering 
    \caption{Customized Deep Sets architecture for PointCloud.}
    \label{tab:deepsets_pointcloud}
    \begin{tabular}{c|c|c}
    \hline
      Encoder & Aggregation & Decoder\\
    \hline
         x - max(x)& Max & Dropout(0.5)\\ 
         FC(256) & & FC(256) \\
         Tanh& & Tanh\\
         x-max(x) & &Dropout(0.5)\\
         FC(256)& &FC(n\_outputs)\\
         Tanh&& \\
         x-max(x)&& \\
         FC(256) & & \\
         Tanh &&\\
        \hline
    \end{tabular}
    \bigskip
      \caption{Customized Set Transformer architecture for PointCloud.}
      \label{tab:settransformer_pointcloud}
    \begin{tabular}{c|c|c}
    \hline
      Encoder & Aggregation & Decoder\\
    \hline
         FC(128) &  Dropout(0.5)  & Dropout(0.5)\\
         ISAB(128, 4, 32) &PMA(128, 4)&FC(n\_outputs)\\
         ISAB(128, 4, 32) &&\\
         \hline
    \end{tabular}
   
\end{table}

\begin{table}[]
    \centering 
    \caption{Point cloud test accuracy}
    \label{tab:pointcloud_acc}
    \begin{tabular}{l|r}
    \toprule
        Model &  Accuracy \\
        \midrule
        Deep Sets &  0.86 \\
        Deep Sets++ & 0.87 \\
        Set Transformer & 0.86\\
        Set Transformer++ & 0.87 \\
        SimpleView & 0.92 \\
        PointNet++ & 0.92\\
        \bottomrule
    \end{tabular}
\end{table}

%% file: appendix/ARC.tex
\section{Aggregated residual connections (ARCs)}
\label{sec:arc}

A function $g$ which adds aggregated equivariant residual connections to any equivariant function $f$ is also permutation equivariant:

\begin{align*}
g(\pi \mbx) &= f(\pi \mbx) + \text{pool}(\mbx_1, \dots, \mbx_S)  \\
&= \pi f(\mbx) + \text{pool}(\mbx_1, \dots, \mbx_S)  = \pi g(\mbx).
\end{align*}

Results in \Cref{tab:type_res} clearly show that clean path ARCs perform worse than clean path ERCs 
(\Cref{tab:clean_nonclean}).

%% file: appendix/residuals_gradients.tex
\section{Gradient Computation for Equivariant Residual Connections}
\label{sec:residual_gradients}
We compute the gradients for early weights in a Deep Sets network with equivariant residual connections below.

We denote a single set $\mbx$ with its samples $\mbx_1, ... \mbx_M$. We denote hidden layer activations as $\mbz_{\ell, i}$ for layer $\ell$ and sample $s$. In the case of no residual connection, $\mbz_{\ell, i} = \text{ReLU}(\mbz_{\ell - 1, m}W_\ell + b_\ell)$. We denote the output after an $L$-layer encoder and permutation invariant aggregation as $\mby = \sum_i \mbz_{L, i}$ (we use sum for illustration but note that our conclusions are the same also for max). For simplicity let the hidden dimension remain constant throughout the encoder.

Now, we can write the gradient of weight matrix of the first layer $W_1$ as follows:
\begin{align}
    \frac{\partial{\mathcal{L}}}{\partial{W_1}} &= \frac{\partial{\mathcal{L}}}{\partial{W_1}} \sum_i \frac{\partial{\mby}}{\partial{\mbz_{L, i}}}
     \frac{\partial{\mbz_{L, i}}}{\partial{W_1}}.
\end{align}

Equivariant residual connections prevent vanishing gradients by passing forward the result of the previous computation along that sample's path, i.e. $\mbz_{\ell, i} = \text{ReLu}(\mbz_{\ell - 1, i}W_\ell + b_\ell) + \mbz_{\ell - 1, i}$:
\begin{align}
    \frac{\partial{\mbz_{L, i}}}{\partial{\mbz_{1, i}}} &=
    \prod_{\ell = 2}^L \frac{\partial{\mbz_{\ell, i}}}{\partial{\mbz_{\ell - 1, i}}}\\
    &= \prod_{\ell = 2}^L \frac{\partial{\text{ReLU}(\mbz_{\ell,i })}}{\partial{\mbz_{\ell, i}}}(1 + W_\ell).
\end{align}